\documentclass[lettersize,journal]{IEEEtran}
\usepackage{amsmath,amsfonts}
\usepackage{algorithmic}
\usepackage{array}
\usepackage[caption=false,font=normalsize,labelfont=sf,textfont=sf]{subfig}
\usepackage{textcomp}
\usepackage{stfloats}
\usepackage{url}
\usepackage{verbatim}
\usepackage{graphicx}
\def\BibTeX{{\rm B\kern-.05em{\sc i\kern-.025em b}\kern-.08em
    T\kern-.1667em\lower.7ex\hbox{E}\kern-.125emX}}
\usepackage{balance}

\usepackage{times}
\usepackage{epsfig}
\usepackage{graphicx}
\usepackage{amsmath}
\usepackage{amssymb}
\usepackage{fancyhdr}

\usepackage{tikz}
\usetikzlibrary{arrows}

\usepackage{color}
\usepackage{xifthen}

\newcommand{\expectation}[2]{\mathbb{E}_{ #2 }\left[#1 \right]}
\newcommand{\parentheses}[1]{\left( #1 \right)}

\newcommand{\tensor}[4]
{
    \ifthenelse {\equal{#2}{}} 
        {\mathbf{#1}^{#3}_{#4}} 
        {\mathbf{#1}\scalebox{.5}{\{$#2$\}}^{#3}_{#4}} 
}
\usepackage[breaklinks=true,bookmarks=false]{hyperref}

\title{Unlocking the potential of two-point cells for energy-efficient and resilient training of deep nets}
\begin{document}
%%%%%%%%% TITLE
\author{Ahsan Adeel$^{1,2,3*}$\thanks{*$^1$Oxford Computational Neuroscience, Nuffield Department of Surgical Sciences, University of Oxford, Oxford. $^2$CMI Lab, University of Wolverhampton, Wolverhampton. $^3$deepCI.org, 20/1 Parkside Terrace, Edinburgh. $^4$ Edinburgh Napier University, Edinburgh. $^5$School of Engineering, University of Edinburgh, Edinburgh. $^6$Department of Psychology, University of Stirling, Stirling. Email: ahsan.adeel@deepci.org}\, Adewale Adetomi$^{2}$\, Khubaib Ahmed$^{2}$\, Amir Hussain$^4$\, Tughrul Arslan$^5$\, W.A. Phillips$^{6}$}

% Unlocking the energy-saving nature of the two-point neuron

%Context-sensitive two-point neuron is inherently energy-efficient

%Context-sensitive two-point neuron for energy-efficient on-chip computing

%Cooperation, not selfishness: Going beyond Loihi, TrueNorth, SpiNNaker 

%Going beyond selfish neurons: Towards  Cooperation, not selfishness: Going beyond point neuron based chip designs

%Cooperation, not selfishness: Going beyond point neuron based chip designs
\maketitle
%\thispagestyle{empty}
%%%%%%%%% ABSTRACT
\begin{abstract}
Context-sensitive two-point layer 5 pyramidal cells (L5PCs) were discovered as long ago as 1999. However, the potential of this discovery to provide useful neural computation has yet to be demonstrated. Here we show for the first time how a transformative L5PCs-driven deep neural network (DNN), termed the multisensory cooperative computing (MCC) architecture, can effectively process large amounts of heterogeneous real-world audio-visual (AV) data, using far less energy compared to best available `point' neuron-driven DNNs. A novel highly-distributed parallel implementation on a Xilinx UltraScale+ MPSoC device estimates energy savings up to 245759 × 50000 $\mu$J (i.e., 62\% less than the baseline model in a semi-supervised learning setup) where a single synapse consumes $8e^{-5}\mu$J. In a supervised learning setup, the energy-saving can potentially reach up to 1250x less (per feedforward transmission) than the baseline model. The significantly reduced neural activity in MCC leads to inherently fast learning and resilience against sudden neural damage. This remarkable performance in pilot experiments demonstrates the embodied neuromorphic intelligence of our proposed cooperative L5PC that receives input from diverse neighbouring neurons as context to amplify the transmission of most salient and relevant information for onward transmission, from overwhelmingly large multimodal information utilised at the early stages of on-chip training. Our proposed approach opens new cross-disciplinary avenues for future on-chip DNN training implementations and posits a radical shift in current neuromorphic computing paradigms.

\end{abstract}

\section{Introduction}
Conventional point neuron \cite{hausser2001synaptic}\cite{burkitt2006review} inspired DNNs have demonstrated ground-breaking performance improvements in a wide range of real-world problems, ranging from image recognition \cite{lecun2015deep} to speech processing \cite{adeel2019lip}\cite{adeel2021contextual}\cite{gogate2020cochleanet}. Scientists have also designed point neuron inspired sophisticated computer architectures e.g., Intel's Loihi \cite{davies2018loihi}, IBM's TrueNorth \cite{merolla2014million},  SpiNNaker \cite{furber2014spinnaker}, Neurogrid \cite{benjamin2014neurogrid}, BrainSclaseS \cite{schmitt2017neuromorphic}, MNIFAT \cite{lichtsteiner2008128}, DYNAP  \cite{moradi2017scalable}, DYNAP-SEL \cite{thakur2018large}, ROLLS \cite{qiao2015reconfigurable}, Spirit \cite{valentian2019fully}, DeepSouth \cite{wang2017neuromorphic}, Tianjic \cite{pei2019towards}, ODIN \cite{frenkel20180}, and Intel SNN chip \cite{chen20184096}. However, point neuron-driven technologies are often economically, technically, and environmentally unsustainable \cite{thompson2020computational}\cite{strubell2019energy}. Their unrealistically high computational demand and complexity scale so rapidly that the technology becomes burdensome \cite{thompson2020computational}. When a single leaky integrate-and-fire (LIF) point neuron fires, it consumes significantly more energy compared to the equivalent computer operation, and an unnecessary fire not only affects the neurons it is directly connected to, but also others operating under the same energy constraint \cite{gangopadhyay2020spiking}. The unnecessary neural firing leads to unnecessary information transmission that creates a huge demand on energy consumption by the system as a whole. Yet, such models can learn, sense and perform complex tasks continuously, but at energy levels that are currently unattainable for modern processors.

\begin{figure*}
    \centering
%    \begin{subfigure}
        \centering
        \includegraphics[width=3in]{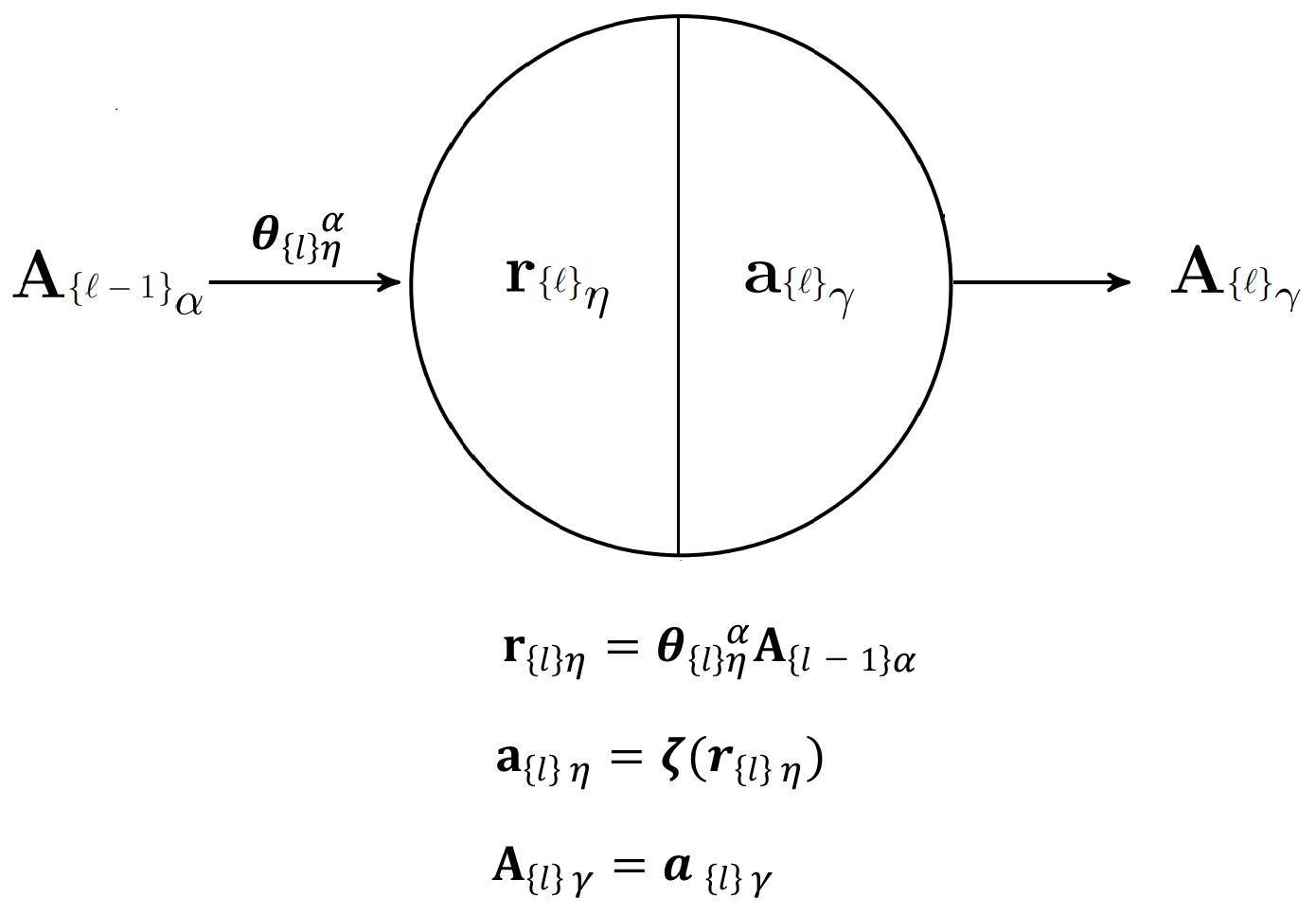}
%        \subcaption{}
%        \caption{}
%    \end{subfigure}%
%    \begin{subfigure}
        %\includegraphics[width=1\textwidth]{Figures/DeepMutualInformation.pdf}
        \includegraphics[width=3.2in]{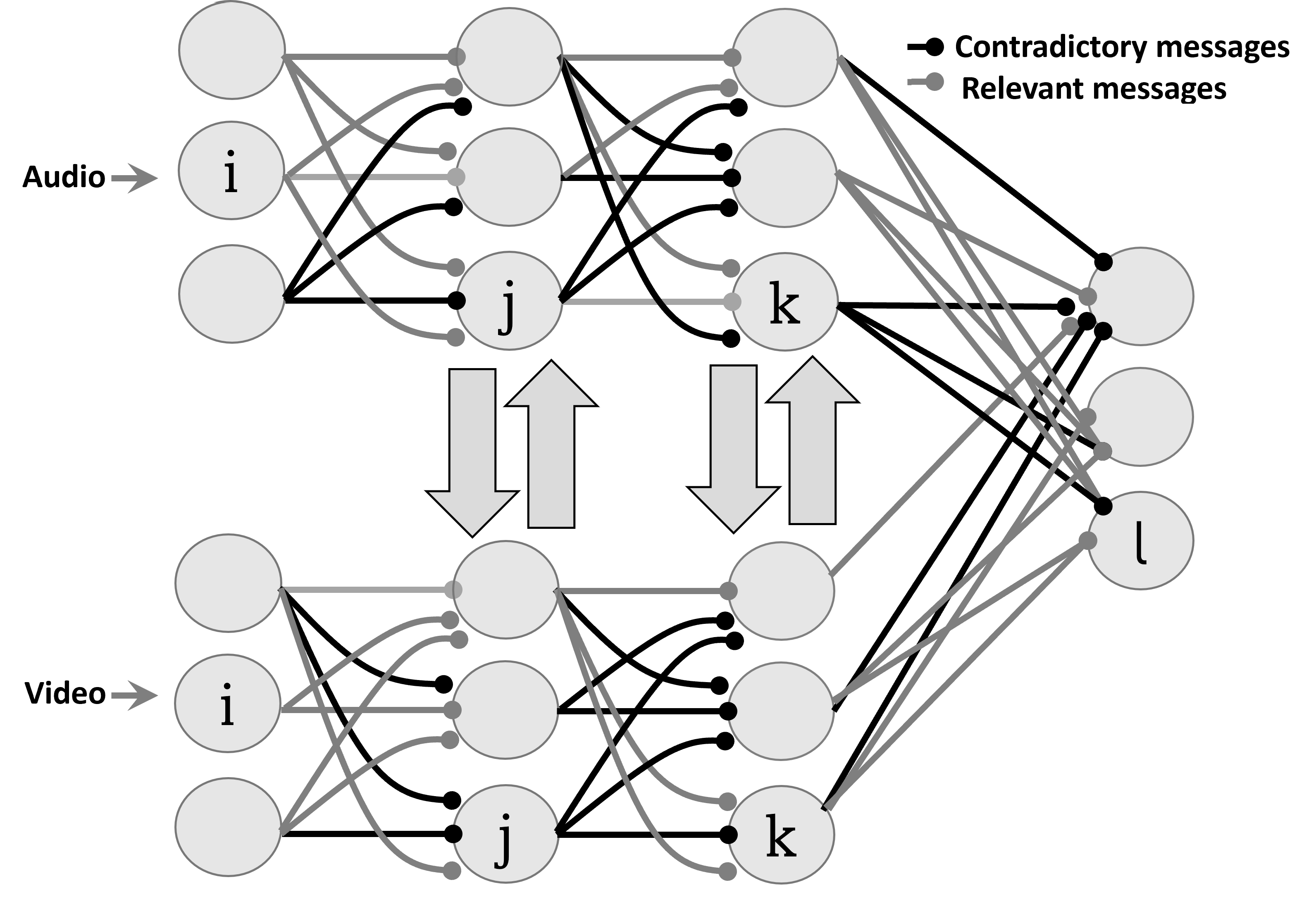}
%        \subcaption{}
%    \end{subfigure}
    \caption{(a) State-of-the-art point neuron (left) \cite{hausser2001synaptic}\cite{burkitt2006review} (b) point neuron based DNN with cross-channel communication (C3)/attention (right)  \cite{yang2019cross}\cite{cangea2019xflow}\cite{guo2019deep}\cite{bhatti2021attentive}. It is to be noted that the point neuron has no inherent mechanism to distinguish between coherent and conflicting messages, hence, it maximises the transmission of every information it receives.}
    \label{fig:deep_mi_task}
\end{figure*}
\begin{figure*}
    \centering
%    \begin{subfigure}
        \centering
        \includegraphics[width=3in]{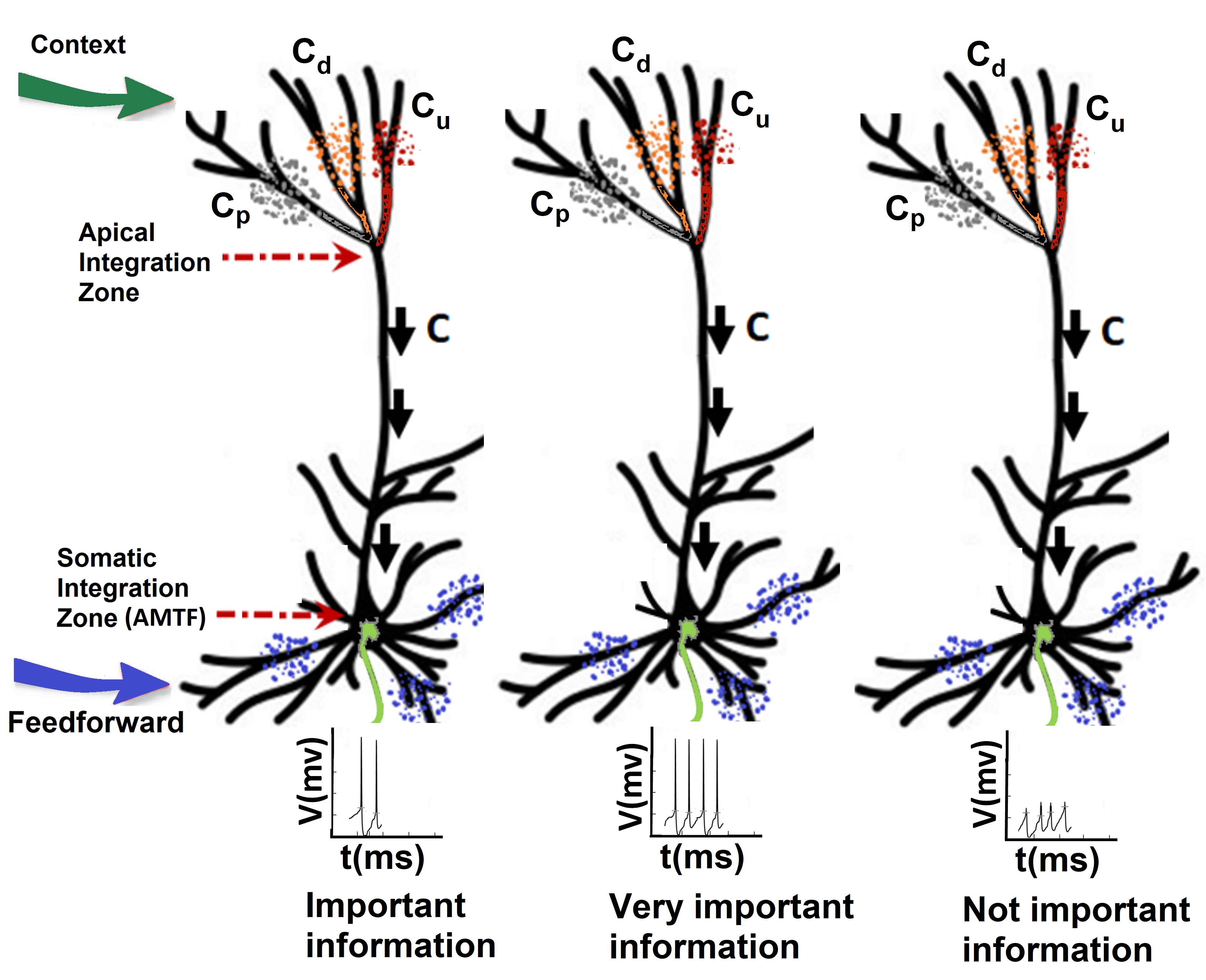}
%        \subcaption{ }
%        \caption{}
%    \end{subfigure}%
%    \begin{subfigure}
        %\includegraphics[width=1\textwidth]{Figures/DeepMutualInformation.pdf}
        \includegraphics[width=3.5in]{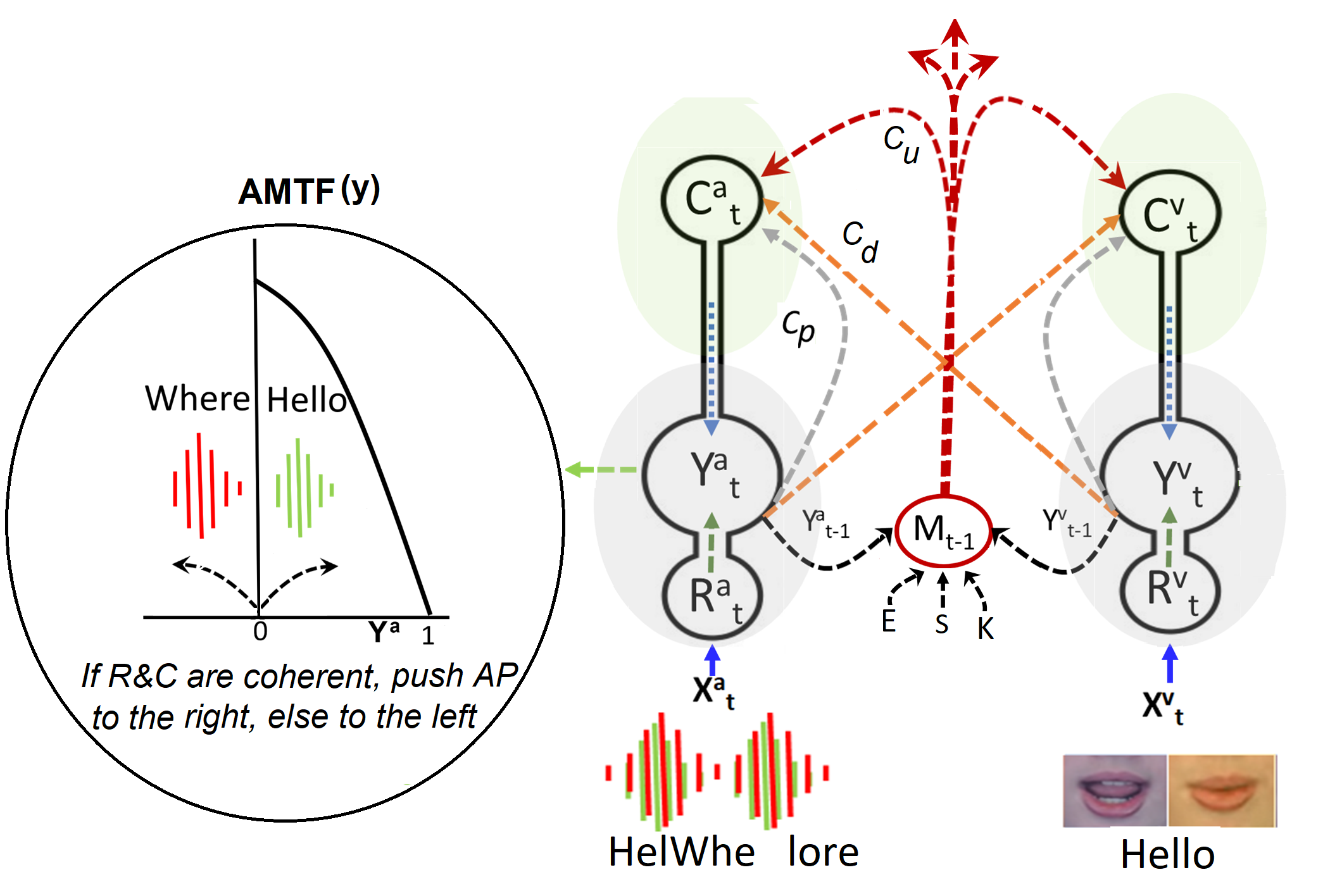}
%        \subcaption{ }
%    \end{subfigure}
    \caption{Multisensory Cooperative Computing \cite{adeel2020multisensory}\cite{adeel2022context}: (a) CMI-inspired \cite{adeel2020conscious} two-point neurons (left) whose apical tuft integrate input from diverse cortical and subcortical sources as a context, including local proximal context ($C_p$), local distal context ($C_d$), and universal context ($C_u$), which are used by the AMTF to decide whether the received information is relevant (important), very relevant (very important), or irrelevant (not important) (b) example L5PC driven AV speech processing (right): the RF (in blue) represents the ambiguous sensory signal (e.g., noisy audio), $C_p$ represents the noisy audio coming from the neighbouring cell of the same network or the prior output of the same cell, $C_d$ represents the signal coming from other parts of the current external input (e.g., visuals), and $C_u$ represents the brief memory broadcasted to other brain regions. The brief memory could explicitly be extended to include prior experiences (E), emotional states (S), and  semantic knowledge (K). The AMTF associated with the audio input splits the coherent and conflicting signals with the conditional probability of Y: $Pr(Y=1|R=r, C=c)=p(T(r,c))$, where \textit{p} is the half-Gaussian filter and T(r,c) is a function defined on $\mathbb{R}^2$.}
    \label{fig:deep_mi_task}
\end{figure*}
The fundamental problem is attributed to the simplified LIF neural structure that processes every piece of information it receives, irrespective of whether or not the information is useful to other neurons or the long-term benefit of the whole network \cite{adeel2022context}. This approach increases the overall neural activity or contradictory messages to high perceptual levels, leading to energy-inefficient and hard to train DNNs \cite{adeel2022context}. Furthermore, the lack of dynamic cooperation between neurons make these DNNs intolerant of faults.  A simple illustration of point neuron and point neuron based neural network is presented in Fig. 1. The point neuron integrates all incoming streams in an identical way i.e., simply summing up all the excitatory and inhibitory inputs, with an assumption that they have the same chance of affecting the neuron's output \cite{hausser2001synaptic}. In contrast, biologically inspired two-point neurons transmit information only when the received information is relevant\footnote{Relevant (coherent) information refers to the portion of input information being logical and consistent with other portions of input information from the source data.} to the task at hand, and not otherwise \cite{adeel2022context}. 

Recent neurobiological breakthroughs \cite{larkum1999new}\cite{larkum2013cellular} have discovered neocortical neurons with two functionally distinct points of integration (apical and basal) in thick-tufted layer 5 pyramidal cells of the mammalian neocortex. However, it has not been demonstrated until now how these cells can provide useful neural computation. Although a few machine learning experts such as G. Hinton \cite{lillicrap2020backpropagation}, T.P. Lillicrap \cite{guerguiev2017towards}, R. Naud \cite{payeur2021burst} and Y. Bengio \cite{sacramento2018dendritic} have been inspired by the discovery of two-point L5PC, their papers have focused predominantly on learning, whereas our work uses context to guide both ongoing processing and learning \cite{adeel2022context}. Guided by the underlying philosophy first espoused in \cite{adeel2022context}\cite{adeel2020conscious}, the main contributions of this paper are as follows:

\begin{itemize}
\item To the best of our knowledge, this study is the first to demonstrate the transformative computational potential of the L5PC for energy-efficient processing of rich real-world multi-modal data for a benchmark AV speech enhancement problem, where multiple real-world noises corrupt speech in real-world like conditions.

\item A novel L5PC inspired context-sensitive cooperative processing unit (CCPU) is proposed that interacts moment-by-moment with other CCPUs in the network, termed MCC, to maximize the transmission of only salient, relevant or coherent activity of the network. Individual CCPUs fire only when the received information is relevant to the task at hand.
 
\item Hardware implementation of our proposed brain-inspired non-von Neumann MCC architecture on a Xilinx UltraScale+ MPSoC device. The hardware architecture emulates the proposed L5PC by not propagating the conflicting messages (represented by the synaptic signal of value zero) in the network, and therefore contributing nothing to the dynamic power consumption. This property is suggested to be very useful for on-chip training and testing of both shallow and DNNs.

\item The proposed method is evaluated with the benchmark AV Grid \cite{cooke2006audio} and ChiME3 \cite{barker2017third} corpora, with 4 different real-world noise types (cafe,
street junction, public transport (BUS), pedestrian area) and compared with popular DNN models for both supervised and unsupervised AV speech processing tasks. Comparative results show that our new method demonstrates superior energy consumption and generalisation performance in all experimental conditions.

\end{itemize}

\begin{figure*}
    \centering
%    \begin{subfigure}
        \centering
        \includegraphics[width=3in]{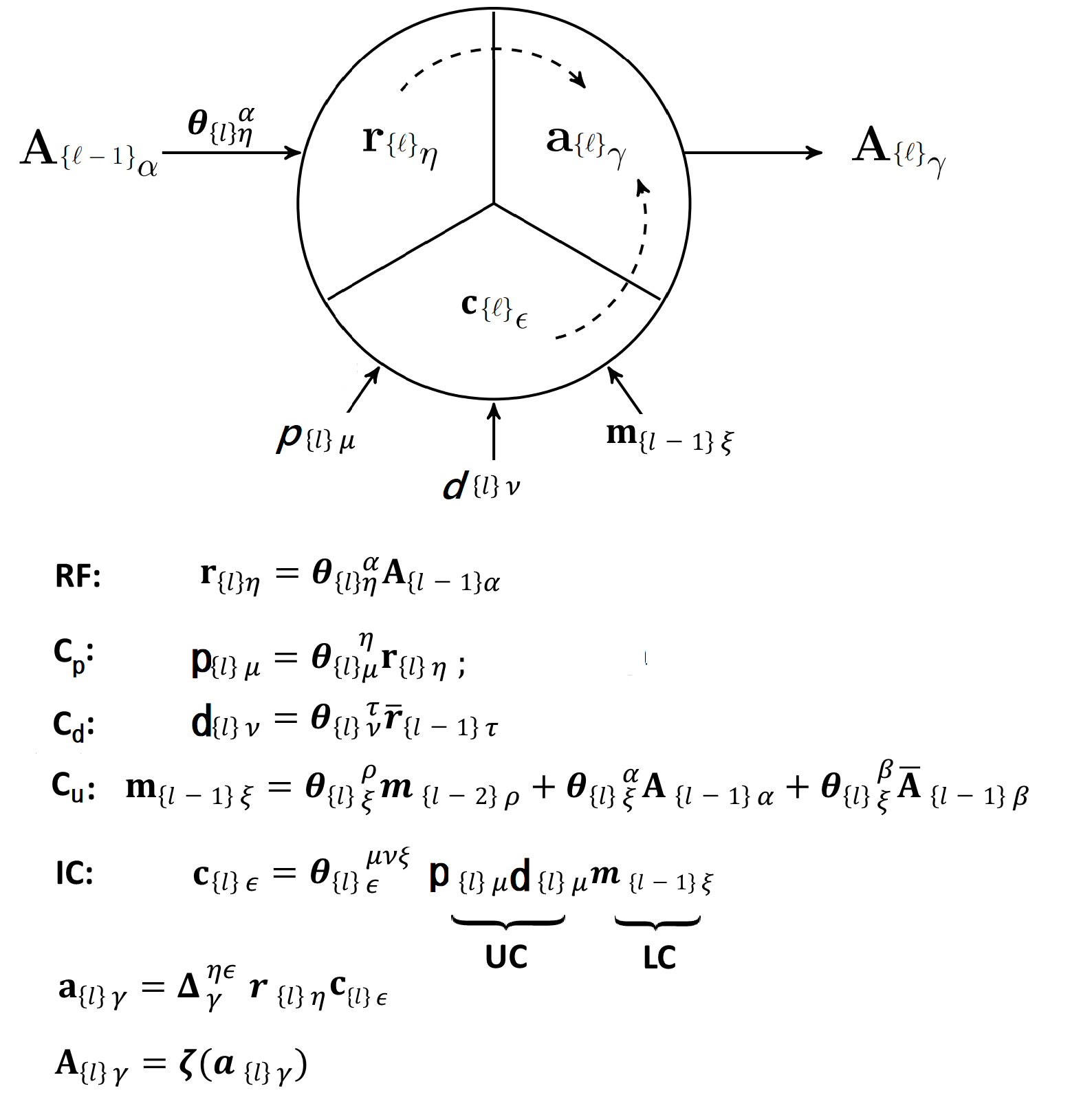}
%        \subcaption{}
%    \end{subfigure}%
%    \begin{subfigure}
        %\includegraphics[width=1\textwidth]{Figures/DeepMutualInformation.pdf}
        \includegraphics[width=3.4in]{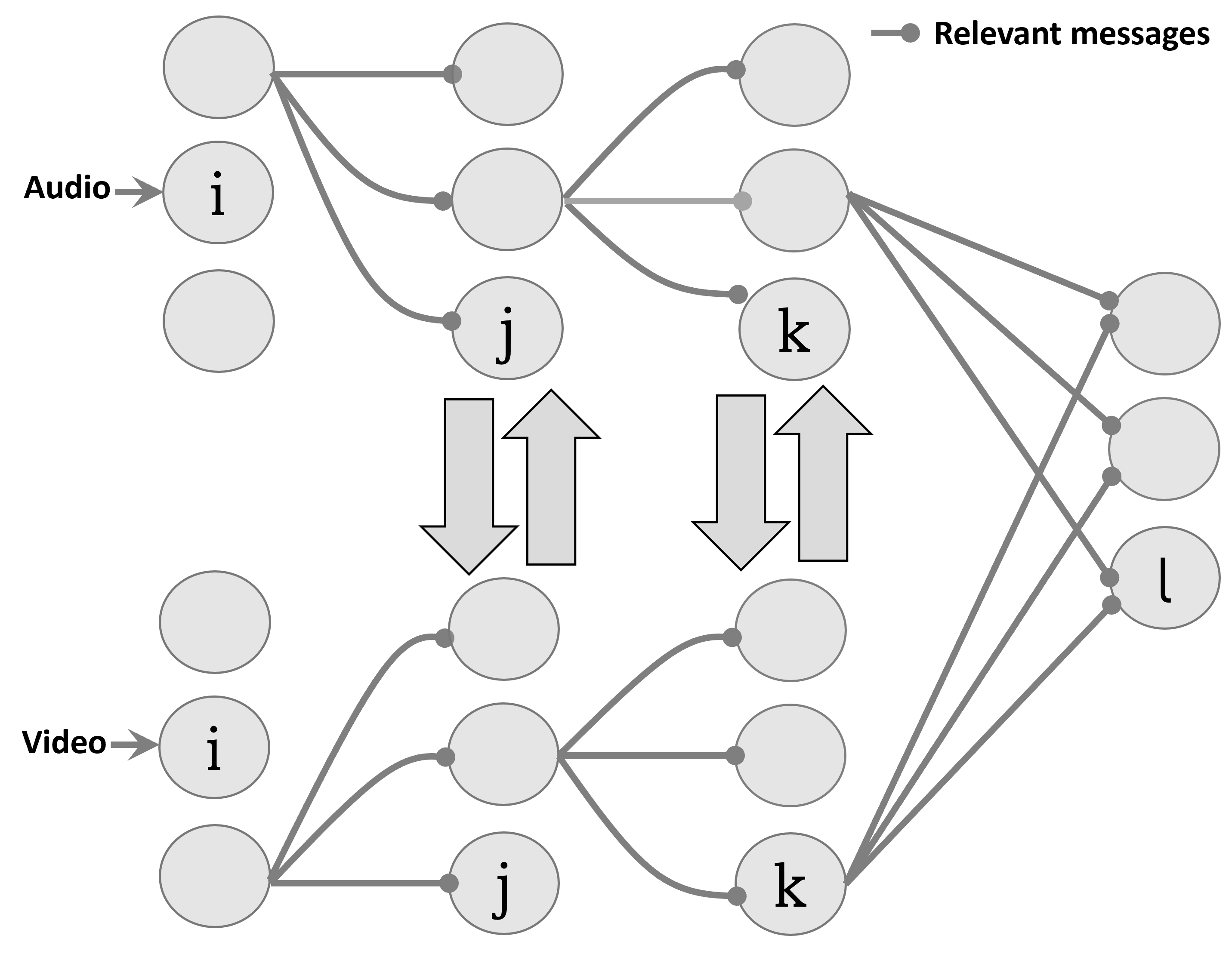}
%        \subcaption{ }
%    \end{subfigure}
    \caption{(a) CCPU (left) comprises: (1) an RF generator (r) configured to generate RF based on inputs to which synaptic weights ($\theta_{{\ell}}{\eta}^{\alpha }$)  are applied (2) an integrated context (C) configured to generate a CF based on inputs to which synaptic weights ($\theta_{{l}\mu}^{\eta}$  $\theta_{{l}\nu}^{\tau}$ $\theta_{{l}\xi}^{\rho}$   $\theta_{\xi}^{\alpha}$ $\theta_{{l}\xi}^{\beta}$ $\theta_{{l}\epsilon}^{\mu \nu \xi}$) are applied (3) an AMTF ($a_{l\gamma }$) configured to generate an output for controlling an activation level of the CCPU based on r and c. The integrated context is dependent on $C_p$, $C_d$, and $C_u$ (b) Multilayered multiunit MCC (right). CCPU in MCC fires only when the received information is coherent across the network or relevant to the task at hand e.g., which data is worth paying attention to and therefore processing just that, instead of having to process everything \cite{adeel2022context}.}
    \label{}
\end{figure*}

\section{Multisensory Cooperative Computing}
In light of conscious multisensory integration (CMI) theory \cite{adeel2020conscious}, Fig. 2 depicts our proposed L5PC that receives three distinct types of contextual fields (CFs) at the apical tuft. These CFs are integrated using a novel 3D-asynchronous modulatory transfer function (3D-AMTF) \cite{adeel2022context}. The 3D-AMTF outputs the conditional probability of Y: $Pr(Y=1|R=r, C=c)=p(T(r,c))$, where \textit{p} is the half-Gaussian filter (HGF) and T(r,c) is a continuous function defined on $\mathbb{R}^2$ and given as $p(R^2 + 2RC + C(1+|R|))$. The modulatory function uses integrated context (C) as a `modulatory force' to push the action potential (Y) to the right or left side of the HGF depending on the relevance or irrelevance of the incoming feedforward information, respectively. This new kind of AMTF goes beyond the conventional contextual modulation \cite{kay2020contextual} and suggests that a strong contextual field (CF) overrules the typical dominance of the RF in deciding whether a particular instance of RF is important, very important, or not important. However, the modulatory function that enables this move systematically could be generated in several different ways, linearly or non-linearly \cite{adeel2022context}\cite{adeel2020multisensory}. This mechanism enables the technical effect of significantly higher energy efficiency and resilience than existing DNN architectures.

Fig. 3a depicts MCC neural model, termed as CCPU. The CCPU interacts with other CCPUs in the network to maximize the transmission of only coherent activity of the network and fires only when the received information is relevant. An example, multiunit two-layered MCC architecture is depicted in Fig. 3(b) and its equivalent hardware model is shown in Fig. 4. The CCPU in one stream is connected to all other CCPUs in adjacent streams of the same layer to effectively coordinate widely distributed and shared activity patterns. This architecture is able to extract synergistic RF components (brief memory, $C_u$) by segregating the coherent and incoherent multisensory information streams  and then recombining only the coherent multi-streams at time \textit{t-1} \cite{aru2020cellular}\cite{bachmann2020dendritic}. The extracted brief memory components $C_u$ are broadcast and received by other CCPUs in the network in their apical tufts at time \textit{t} along with the current local contexts $C_p$ and $C_d$. $C_u$, $C_p$, and $C_d$ are summed to construct an integrated context (IC) represented as C using a simple adder and a non-linear activation function. At time \textit{t-1}, the CF only comprises the external context (i.e., local context) e.g., processed visual streams at the audio channel which modulates the RF using the modulatory transfer function (transfer circuit). The extracted coherent RF signals are then fed into a cross-modal working memory to extract the synergistic components (i.e., universal context). The universal context at time t is combined with the local context to form the integrated context (C) which modulates (amplify or attenuate) the cell’s responses to the feedforward RF input. 
\subsection{Hardware Architecture}
\begin{figure*}
    \centering
%    \begin{subfigure}
        \centering
        \includegraphics[width=3in]{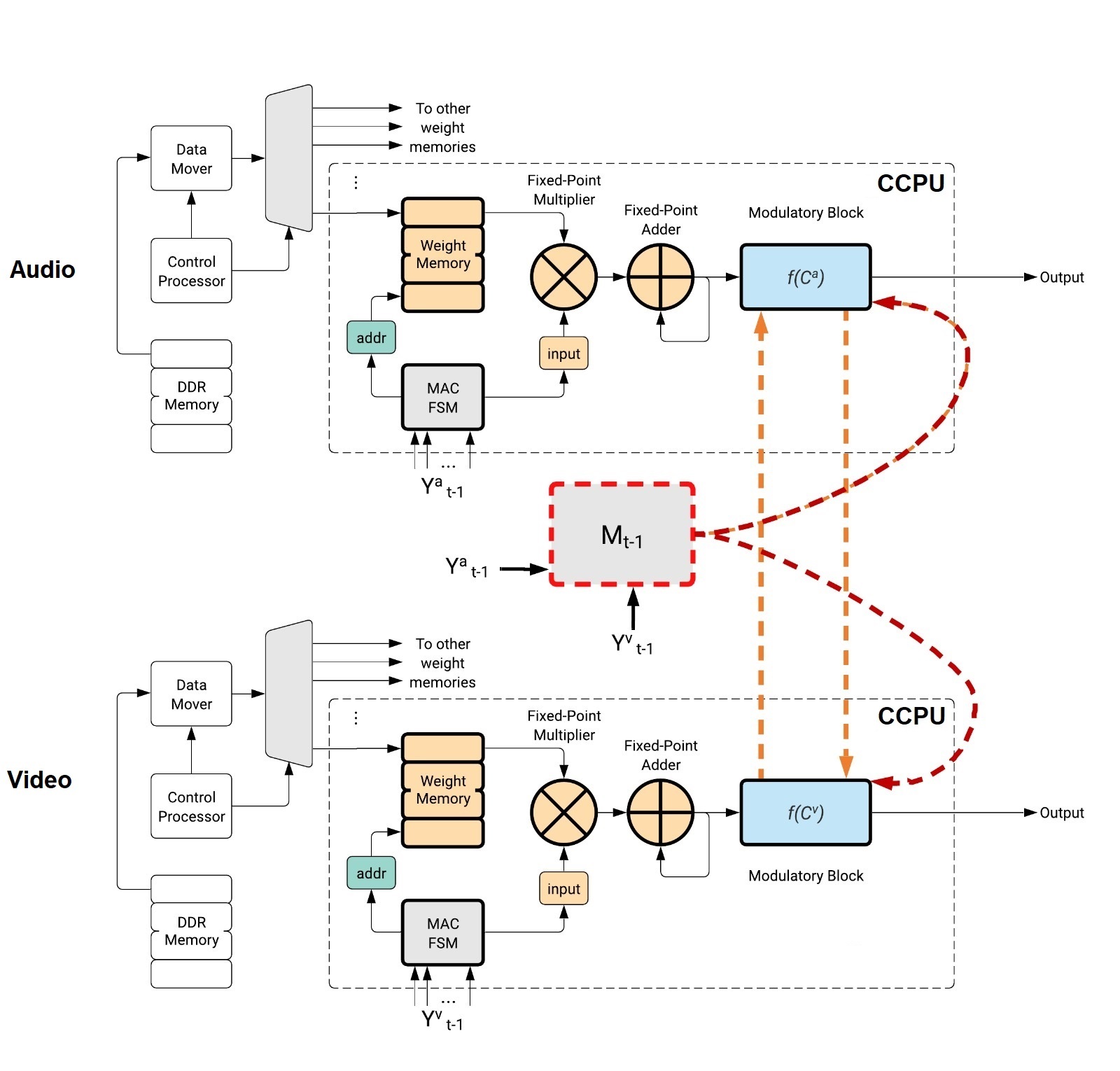}
%        \subcaption{}
%    \end{subfigure}%
%    \begin{subfigure}
        %\includegraphics[width=1\textwidth]{Figures/DeepMutualInformation.pdf}
        \includegraphics[width=3.5in]{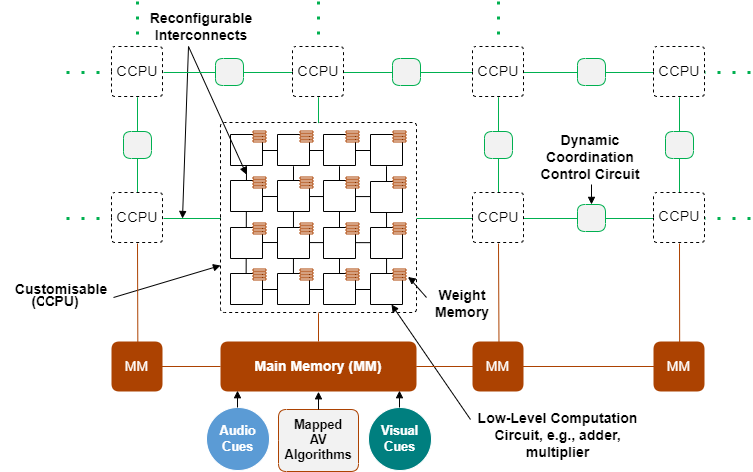}
%        \subcaption{ }
%    \end{subfigure}
    \caption{Hardware architecture: (a) two-CCPU circuit (left) (b) MCC system level architecture (right).}
    \label{fig:deep_mi_task}
\end{figure*}
To emulate the CCPU behaviour and estimate the energy consumption, the hardware architecture is designed such that a synaptic signal of value zero does not propagate in the network and contributes nothing to the dynamic power consumption because of no switching activity. The energy saving per zero-signal synapse per single feed-forward propagation is used to estimate the shallow and deep models energy consumption. For prototyping, the Xilinx UltraScale+ MPSoC device has been targeted. This device offers a high number of configurable logic blocks and block memories needed to implement the proposed architecture. Specifically, we have implemented the prototype on the Genesys ZU-3EG board: xczu3eg-sfvc784-1-e UltraScale+ MPSoC chip. Fig. 4(a) depicts the key building blocks used to build CCPU hardware architecture, including: data loading mechanism, multiply-accumulate finite state machine, weight memory, multiplier, adder, modulatory and activation block. \\
Fig. 4(b) is a diagrammatic representation of the proposed hardware-based MCC system-level architecture. The idea is to replace the neuron with a customizable CCPU, which contains a grid of interconnected computation circuits (e.g., adders and multipliers), where the functionality of the processing unit can be reconfigured on the fly. An array of these customisable CCPUs is linked together by reconfigurable interconnects, such that multiple CCPUs can be dynamically interconnected for distributed processing. With the possibility for the network architecture to change in real-time, the advantages include reduced system downtime, high throughput, and improved robustness. Inter-CCPU coordination control circuits can be implemented to achieve the dynamic behaviour. These control circuits are distributed in the network and can serve as a bridge between CCPUs. Memory bottleneck can be reduced by having the outputs of CCPUs in one layer routed directly to CCPUs in another layer through the coordination control circuits. Multiple high-bandwidth memory blocks can be used to feed in data into the network. The memory interface can be implemented in part with the high-performance ports available in the Xilinx APSoC and MPSoC devices.\\
\textbf{Highly Distributed Parallel Implementation:} The hardware implementation is based on a massively parallel architecture, where each CCPU in every layer has a digital signal processor (DSP) engine that computes the products of the respective features and weights and the sums of these products. A key enabler is the use of local weight memories, implemented with on-chip block memories and attached to the CCPU to remove the bottleneck of memory transfer during execution. Each CCPU in every layer is physically implemented, and all the CCPUs computed in parallel. The key advantage of this parallel approach is the reduction in computation latency, though at the cost of increased resource utilization. Another advantage of the distributed architecture is that it is more readily amenable to regression or classification tasks. A bottom-up approach to the implementation has been adopted, where higher-level components (e.g., multiply-accumulate block) are implemented from lower-level modules (e.g., fixed-point adders and multipliers). The lower-level component themselves are built from device-independent logic and memory elements as much as possible, such that the architecture can be easily ported to other FPGA families and manufacturers. The overarching effect of this is that many different network prototypes of varying complexity can be relatively easily implemented, limited only by the size of the FPGA. Where FPGA size becomes a limitation and the next available FPPGA is out of reach (due to lower power requirements), then an iterative architecture with a single layer can be used to successively compute all the layers. This would imply a reduction in parallelism and a substantial increase in latency, as weights and inputs would have to be loaded for each layer in turn. \\
\textbf{Data Representation:} The number format adopted is 16-bit signed 2's complement fixed-point representation. This is Q3,12 in the Q notation and implies 1 sign bit, 3 integer bits, and 12 fractional bits; with maximum representable integer part of 7, maximum representable number of 7.999755859375 (0x7FFF in hexadecimal format), lowest representable number of -8.0 (0x8000), and a precision of 0.000244140625 (0x0001).\\
\textbf{Bias Modelling}: The bias for each neuron is modelled as an input of 1 and a weight representing the bias. The input of 1 for the bias is hard-coded. The corresponding weight representing the bias is kept in the Weight Memory in the next memory location following all the network weights. In the MAC computation inside each neuron, once all the respective inputs and weights have been multiplied and accumulated, the bias is retrieved from the Weight Memory and added to the accumulated result.\\
\textbf{Weight Memory:} This keeps all the synaptic weights for all the connections feeding the neuron. In addition, it holds the bias. It is implemented with a Block RAM (BRAM) with support for up to 1023 weights and one bias.\\
\textbf{Fixed-Point Multiplier:} This is the implementation of a fixed-point multiplier, taking on the input interface, two 16-bit signals and outputting a 16-bit result and an overflow flag. For the 16-bit inputs, an output of 32-bit would be expected. However, because of the need to maintain the 16-bit data path across the network, the result is quantized by taking the upper 16 bits of the resulting product.\\
\textbf{Fixed-Point Adder:} The fixed-point adder takes two 16-bit inputs and produced a 16-bit result, all in Q3,12 fixed-point format. This module is also used for subtraction. Since the input signals are represented in signed 2’s complement format, a subtraction is essentially an addition. That is, X-Y = X+Y*, where Y* is the 2’s complement of Y. The implementation ensures that arithmetic overflow is detected and addressed. An overflow has occurred if the sum of two positive numbers yields a negative result, or if the sum of two negative numbers yields a positive result. In the former case, we set the result to the maximum number representable in the chosen Q3,12 data format, which is 7.999755859375, or 0x7FFF in the hexadecimal format; while in the latter case, we set the result to -8.0, or 0x8000.\\
\textbf{Modulatory function:} This is the implementation of the modulatory function (Mod), requiring the use of the fixed-point multiplier and fixed-point adder. For improved efficiency in the resulting implemented hardware, we write the function as $Mod=p(2R^2 + R+R + 2C(1+|R|))$, where \textit{p} is the Relu6. Addition is less computationally intensive and more resource-efficient than multiplication. One multiplication and one addition are more efficient than two multiplications. We
re-arranged the original equation to reduce the computation complexity. The rearranged equation requires two multipliers and two adders against three
multipliers and three adders for the original equation.\\
\textbf{Activation Block:} The Activation Block implements the required activation function (ReLU). It is a parameterized block that includes only the required activation function at compile time. The ReLU implementation in hardware is straightforward, producing an output zero if the input is less than zero, and raw output otherwise. The maximum positive output is also clipped to a value of 6. \\
\textbf{MAC FSM:} The Multiply-Accumulate Finite State Machine (MAC FSM) controls the retrieval of the weights from the Weight Memory and the selection of the corresponding inputs. It automatically advances the address value fed into the memory every clock cycle.\\
\textbf{Data Loading Mechanism:}
This comprises the Control Processor, the Data Mover, and the DDR Memory. The Control Processor is an Arm Cortex-A53 processor in the UltraScale+ MPSoC device, used for loading the weights and the inputs from an eternal DDR memory into the programmable logic of the FPGA. A \textit{C} application running on bare-metal OS has been used for prototyping. Routines were written to mount a micro-SD card, from where the weights and inputs are transferred to the DDR Memory using the Data Mover which is a direct memory transfer engine. Thousands Weight Memories are required to be filled with weights and biases. A multiplexing approach is taken for this data loading, where the Weight Memories are attached in turn to the Data Mover for all their respective weights and biases to be loaded. This is a step that happens once after power up and does not impact on the latency of inference. A similar multiplexing solution is adopted for loading the inputs for the input layer. \\
\textbf{Feasibility to Implement on Integrated Circuits and Chips:} FPGAs typically excel as a viable means of design and verification of hardware-based functionality before committing to fixed silicon (ASIC), thanks to their programmability. As such, the architecture being implemented aligns well with this paradigm. The synthesis and implementation artifacts of the hardware build process are standardized outputs that can be passed on to the ASIC fabrication process.\\

\section{Experiments}
The ability of the proposed method is demonstrated and compared with sophisticated and popular shallow and deep learning approaches \cite{belghazi2018mutual}\cite{cangea2019xflow}\cite{yang2019cross}\cite{guo2019deep}\cite{bhatti2021attentive} on a challenging noisy audio-visual speech processing task that uses video information from lip movements to selectively amplify speech signals heard in noisy environments. It is observed that MCC is able to remove background noise with better reconstruction than the state-of-the-art baseline shallow and deep learning algorithms. For fair comparisons, both shallow and deep benchmark models have C3/attention blocks integrated \cite{yang2019cross}\cite{cangea2019xflow}\cite{guo2019deep}\cite{bhatti2021attentive}. The C3 or cross-channel fusion is implemented through concatenation, addition, or multiplication using LIF-inspired point neural model. All models have a similar structure and similar layers between different models and have the same configuration. For testing, the Grid \cite{cooke2006audio} and ChiME3 \cite{barker2017third} corpora are used \cite{adeel2021contextual}, including four different noise types; cafe, street junction, public transport, and pedestrian area with the signal-to-noise ratio (SNRs) ranging from -12dB to 12dB with a step size of 3dB. For shallow models, logFB audio features of dimension 22 and DCT visual features of dimension 50 were used \cite{adeel2019lip}. The shallow baselines include popular mutual information neural estimation (MINE) approach \cite{belghazi2018mutual}, state-of-the-art concatenation approach \cite{Poria2017}, and cross-modal approach \cite{bhatti2021attentive}. The shallow models pose the problem of semi-supervised AV speech processing with the following loss function:

\begin{align*}
    \mathcal{L}_{1} &= \beta \expectation{\text{SE}\parentheses{\tensor{Z}{}{}{},\tensor{\hat{Z}}{}{}{}}}{} -\alpha\expectation{-I_{f}\parentheses{\tensor{X_\alpha}{}{}{} ; \tensor{Y_\beta}{}{}{} }}{} 
\end{align*}

where the first term in the equation above is the squared error (SE) between the clean target speech (Z) and clean predicted speech ($\hat{Z}$). The second term represents the mutual information (MI) between audio ($X_\alpha$) and video ($Y_\beta$) \cite{belghazi2018mutual}. 

For deep models, we used the following loss function:

\begin{figure*}
    \centering
%    \begin{subfigure}
        \centering
        \includegraphics[width=3.64in]{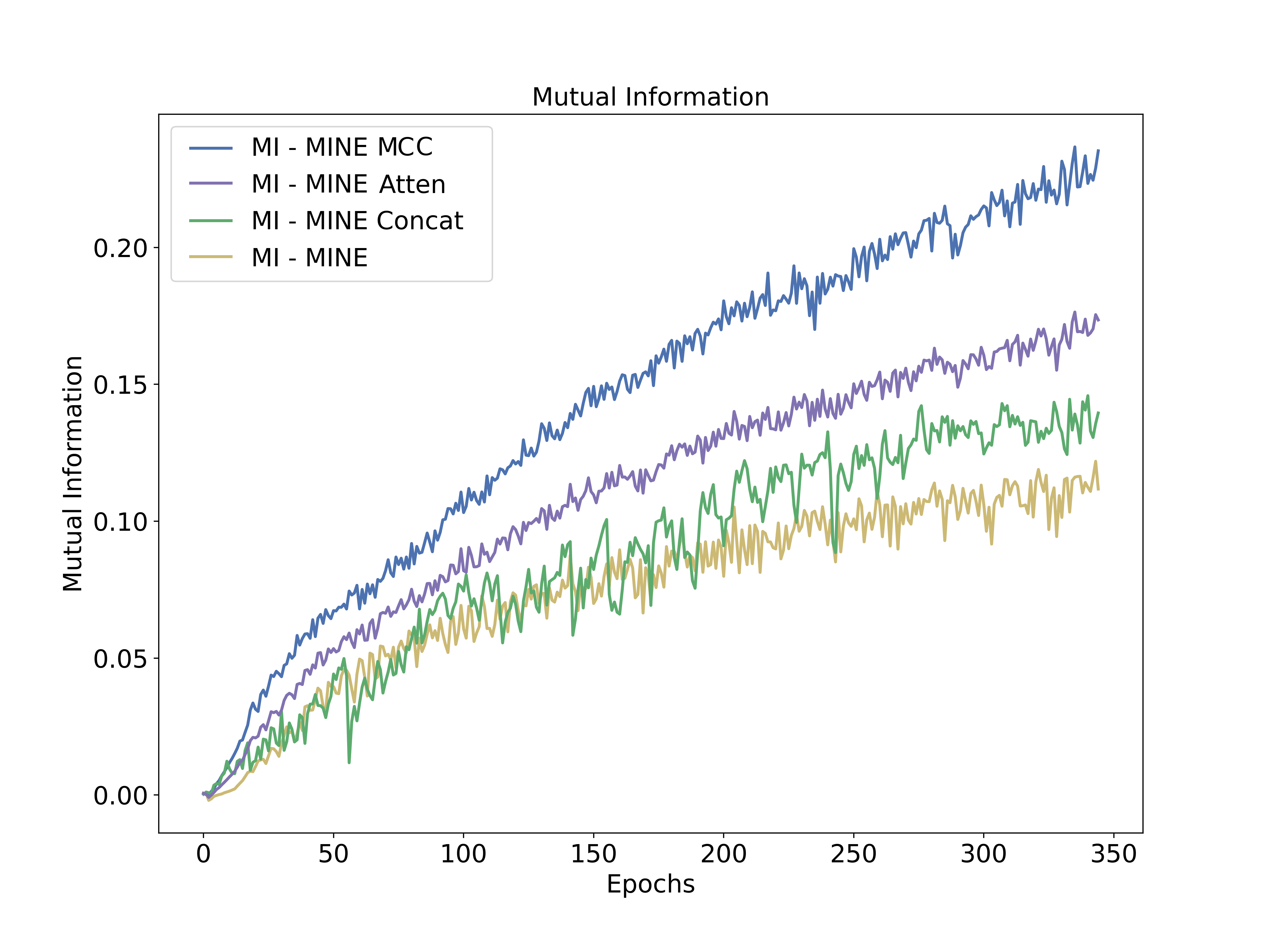}
%        \subcaption{}
%    \end{subfigure}%
%    \begin{subfigure}
        %\includegraphics[width=1\textwidth]{Figures/DeepMutualInformation.pdf}
        \includegraphics[width=3.3in]{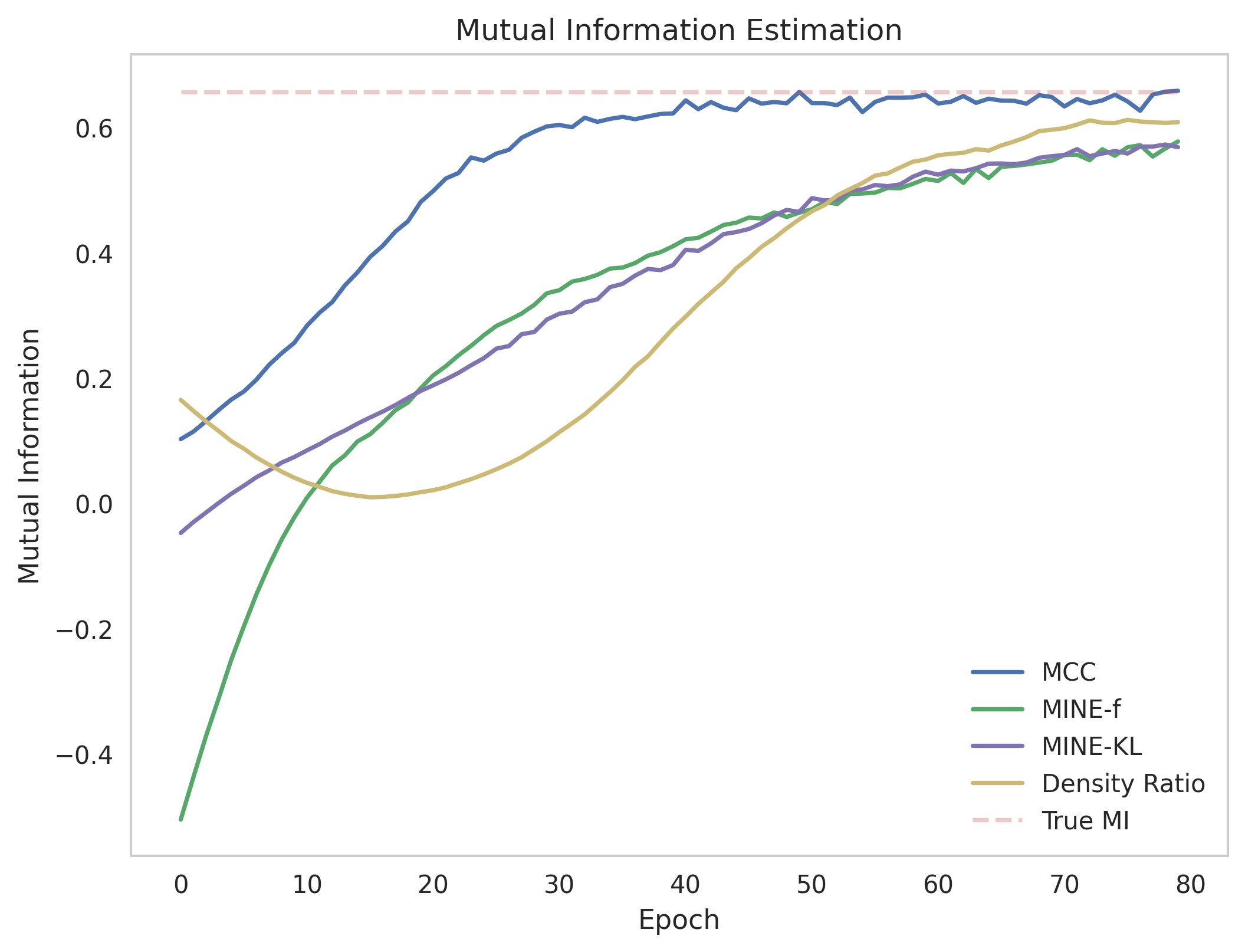}
%        \subcaption{}
%    \end{subfigure}
    \caption{Shallow models: semi-supervised AV training via MI maximization (left) (b) empirical evaluation and comparisons \cite{belghazi2018mutual} (right).}
    \label{fig:deep_mi_task}
\end{figure*}

\begin{table*}[]
\footnotesize
    \centering
    \caption{Shallow models: testing MSE, MAC operations, energy consumption, and latency.}
\begin{tabular}{|l|l|l|l|l|}
\hline
                     & MINE \cite{belghazi2018mutual} & MINE Concat \cite{Poria2017} \cite{kose2021multimodal}  & MINE Attention (Baseline) \cite{bhatti2021attentive} & MCC \\ \hline
Minimum MSE (quantized/unquantized)          & 0.07/ 0.104      & 0.05/ 0.091                     & 0.03/ 0.066                          & 0.018/ 0.039   \\ \hline
Trainable Parameters & 10685 & 16331                        & 26723                         & 10685   \\ \hline
Cells not Firing      & 48.6\%       &61\%                   & 51\%                          & 80\%      \\ \hline
MAC (total/ used)& 10200/ 5243    & 25036/ 9765   & 19432/ 9522    & 10480/ 2306   \\ \hline

Energy  ($\mu$ J) &     0.418     &   0.781  &  0.761    & 0.184     \\ \hline
Latency ($\mu$ s) &     2.25      &   4.28    & 5.52   & 1.60   \\ \hline

\end{tabular}
\end{table*}
\begin{table}[]
    \footnotesize
    \centering
    \caption{Shallow models: resource utilisation. CLB, LUT, and RAMB stands for configurable logic block, look-up table, and random access memory block, respectively.}
\begin{tabular}{|l|l|l|l|}
\hline

                &           & {\textbf{MCC}} &  {\textbf{Baseline}}  \\ \hline
Resouces        & Available   & \% Utilisation                &   \% Utilisation              \\ \hline
CLB             & 8820        &     74.37\%                  &   98.79\%                     \\ \hline
LUT as Logic    & 70560      &     44.57\%                  &   81.45\%                     \\ \hline
LUT as Memory   & 28800        &     2.58\%                    &   2.58\%                      \\ \hline
CLB Registers   & 141120    &     15.58\%                &  26.78\%                     \\ \hline
RAMB18          & 432          &     24.54\%                  &   60.65\%                     \\ \hline
DSP48           & 360          &     34.22\%                 &   72.78\%                     \\ \hline
\end{tabular}
\label{table:resources}
\end{table}

\begin{align*}
    \mathcal{L}_{2} &= \beta \expectation{\text{SE}\parentheses{\tensor{Z}{}{}{},\tensor{\hat{Z}}{}{}{}}}{} + \gamma\expectation{\mathcal{E}}{}
\end{align*}
$\mathcal{E}$ is a differentiable approximation for the number of firings. We adjust the coefficients of the loss functions to make the secondary objectives significantly less important than the main goal; in particular, we set $\gamma$ to a really small value in all experiments. For deep learning, the input was a tuple containing a noisy audio short-time fourier transform (STFT) of dimension 64X64 and a snapshot of the lip movement of dimension 88X44. The output was a clean audio signal (STFT) of dimension 64X64 \cite{adeel2019lip}\cite{adeel2021contextual}. The training and testing split was 80:20. All data is normalized across the whole dataset and presorted to break all order correlations. 
%The dataset is shuffled once more with a seed to add some variability between different runs and to ensure that different models encounter a similar landscape. 

\section{Results}
For energy consumption estimations, a synaptic value of zero from a preceding layer fed into a subsequent layer is taken to contribute nothing to the energy consumption, since a zero input implies no switching activity. The MAC unit computes the product of an input and its corresponding weight and accumulates this result in only 4 clock cycles. The dynamic power consumption of the MAC unit is 2 mW as reported by the XPower Estimator tool. This power is contributed wholly by the fixed-point multiplier. The fixed-point adder is a purely combinational implementation, and as such, has no dynamic power component. At the prototype frequency of 100 MHz, the energy consumption due to an activated neuron that propagates output through its associated synapses to another neuron is therefore equivalent to 2 mW X 4 clock cycles X 10-ns period, which is equal to 0.08 nJ per synapse in a single inference run. This implies that when a neuron is not firing, each associated synapse does not propagate the zero signal and therefore, saves 0.08 nJ per single inference run. To calculate latency, the networks were implemented and run at a clock frequency of 100 MHz. It is pertinent to state that at this speed, no timing error has been observed because the interface port signals are registered to break long combinational paths. 

\begin{figure*}
    \centering
%    \begin{subfigure}
        \centering
        \includegraphics[width=3.5in]{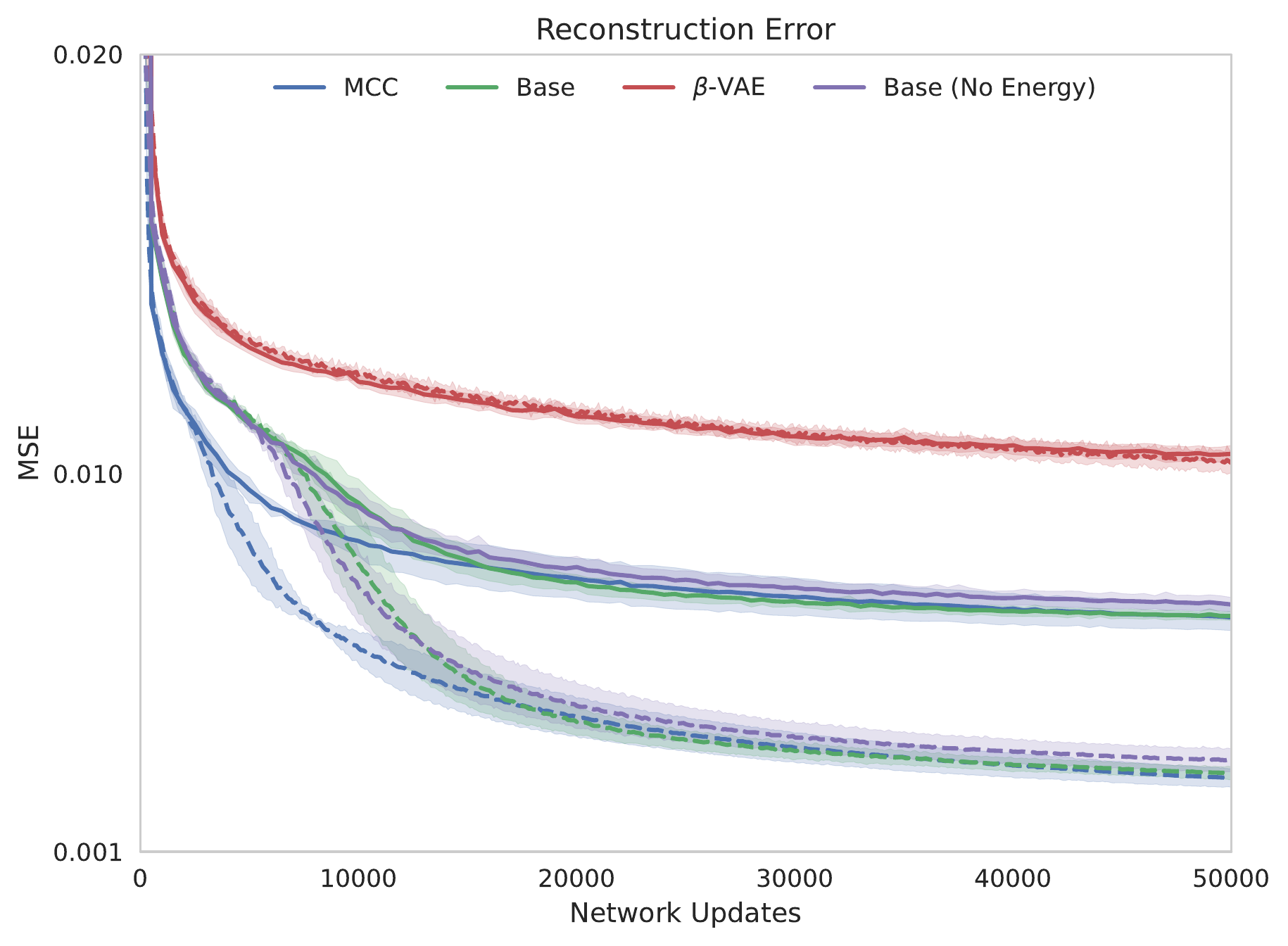}
%        \subcaption{}
%    \end{subfigure}%
%    \begin{subfigure}
        %\includegraphics[width=1\textwidth]{Figures/DeepMutualInformation.pdf}
        \includegraphics[width=3.5in]{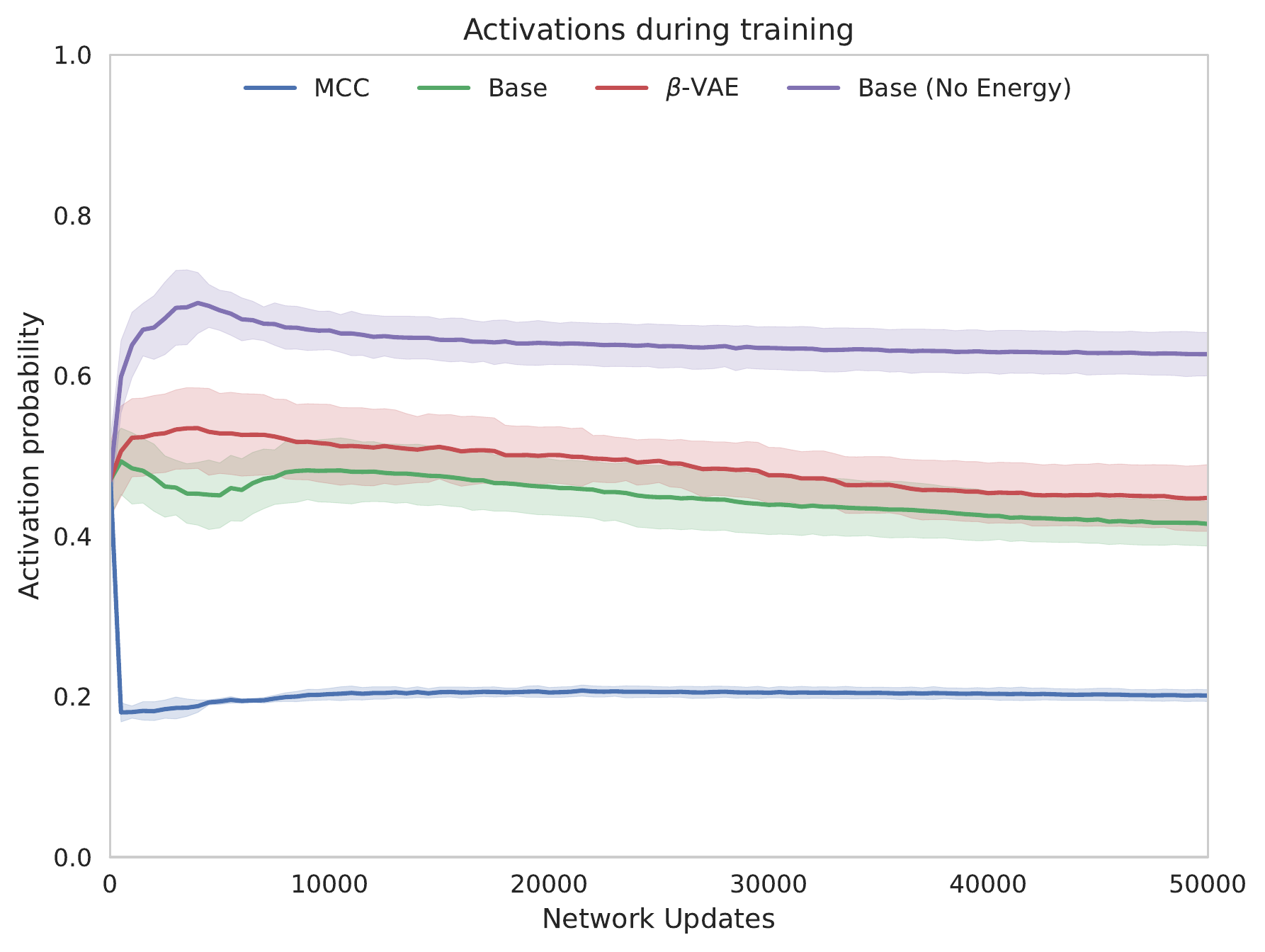}
%        \subcaption{}
%    \end{subfigure}
        \caption{Deep MCC (a) reconstruction error (left) (b) firing evolution (right). MCC learns significantly faster as compared to the the baseline with only 20\% neural activity. Note that the neurons in MCC quickly evolve to become highly sensitive to relevant information and become active (or fire) only when the received information is important for the task at hand. This reduces the overall neural activity and suppresses the transmission of contradictory messages to higher perceptual levels. Solid and dashed lines indicate testing loss and training loss, respectively.}
    \label{fig:deep_multimodal_representation}
\end{figure*}

\begin{figure*} [htb!]
\centering
	\includegraphics[trim=0cm 0cm 0cm 0cm, clip=true, width=0.85\textwidth]{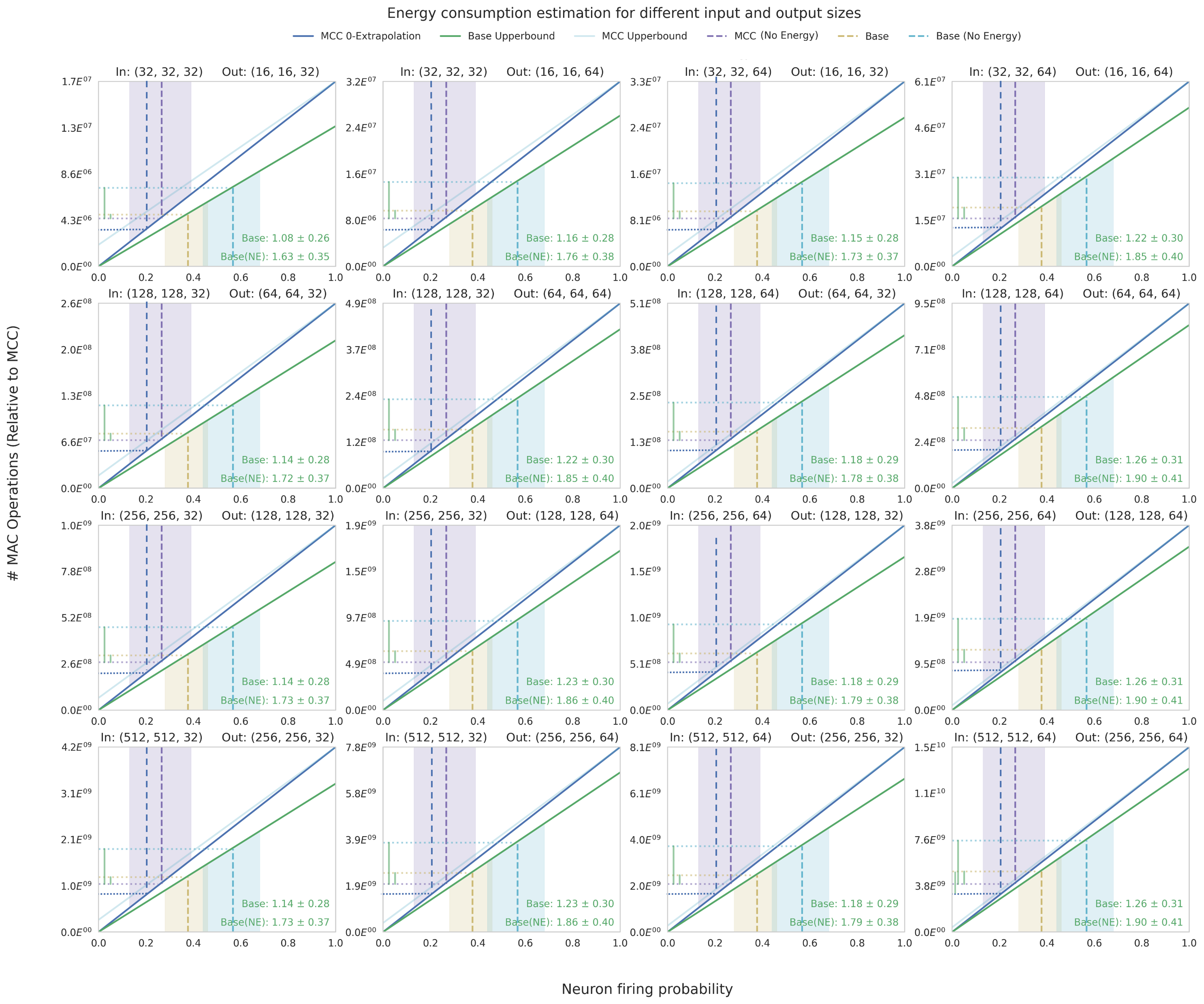}
	\caption{Deep MCC: MAC operations for different number of inputs and outputs. Here a variety of CNN layers are considered for analysis. For a standard CNN model, input is a 3D array with the width, height of a feature map, and the number of feature maps. Similarly, the output is the dimensions of output feature maps.}
	\label{fig:Picture2}
\end{figure*}

\begin{table*}[]
%\tiny
\footnotesize
\centering
\caption{Deep MCC estimated energy consumption per inference without sparsity.}
\begin{tabular}{|l|l|l|l|l|l|l|l|l|l|l|l|l|}
\hline
            & \multicolumn{3}{c|}{\textbf{Input / Output}}  & \multicolumn{3}{c|}{\textbf{Input / Output}}  & \multicolumn{3}{c|}{\textbf{Input / Output}}  \\ \hline
            & \multicolumn{3}{c|}{\textbf{32, 32, 32 / 16, 16, 32}} & \multicolumn{3}{c|}{\textbf{32, 32, 32 / 16, 16, 64}} & \multicolumn{3}{c|}{\textbf{32, 32, 64 / 16, 16, 32}} \\ \hline
            
            & \textbf{MCC}  & \textbf{Baseline}    & \textbf{Saving}    &   \textbf{MCC}    &  \textbf{Baseline}   &   \textbf{Saving} &   \textbf{MCC}    &   \textbf{Baseline}   &   \textbf{Saving} \\ \hline
\textbf{MAC}    & 6268 k     & 9699 k     &	3431 k    &   6348 k      &   9699 k     &   3351 k      &   11904 k   &   19399 k     &   7495 k\\ \hline
\textbf{Energy} & 501 µJ     & 776 µJ     &   274 µJ    &   508 µJ   &   776 µJ  &   268 µJ    &   952 µJ   &   1552 µJ     &   600 µJ\\ \hline

            & \multicolumn{3}{c|}{\textbf{32, 32, 64 / 16, 16, 64}} & \multicolumn{3}{c|}{\textbf{128, 128, 32 / 64, 64, 32}} & \multicolumn{3}{c|}{\textbf{128, 128, 32 / 64, 64, 64}} \\ \hline
            
%                & MCC      &   Baseline    &   Saving      &   MCC         &   Baseline    &   Saving      &   MCC         &   Baseline    &   Saving\\ \hline
\textbf{MAC}    & 51211 k     &   77595 k     &   26384 k      &   95118 k    &   155109 k    &   59991 k   &   98792 k    &   155189 k    &   56397 k\\ \hline
\textbf{Energy} & 4097 µJ     &   6208 µJ     &   2111 µJ      &   7609 µJ    &   12409 µJ    &   4799 µJ   &   7903 µJ    &   12415 µJ    &   4512 µJ\\ \hline

            & \multicolumn{3}{c|}{\textbf{128, 128, 64 / 64, 64, 32}} & \multicolumn{3}{c|}{\textbf{128, 128, 64 / 64, 64, 64}} & \multicolumn{3}{c|}{\textbf{256, 256, 32 / 128, 128, 32}} \\ \hline
            
%               & MCC           &   Baseline    &   Saving  &   MCC         &   Baseline    &   Saving  &   MCC         &   Baseline    &   Saving \\ \hline
\textbf{MAC}    & 185853 k      &   310378 k    &   124526 k &   204044 k    &   310378 k    &   106334 k &   379437 k  &   620757 k    &   241320 k\\ \hline
\textbf{Energy} & 14868 µJ      &   24830 µJ    &   9962 µJ &   16324 µJ    &   24830 µJ    &   8507 µJ &   30355 µJ  &   49661 µJ    &   19306 µJ \\ \hline

            & \multicolumn{3}{c|}{\textbf{256, 256, 32 / 128, 128, 64}} & \multicolumn{3}{c|}{\textbf{256, 256, 64 / 128, 128, 32}} & \multicolumn{3}{c|}{\textbf{256, 256, 64 / 128, 128, 64}} \\ \hline
            
%               & MCC       &   Baseline    &   Saving      &   MCC         &   Baseline    &   Saving      &   MCC         &   Baseline    &   Saving \\ \hline
\textbf{MAC}    & 394612 k  &   620757 k    &   226145 k     &   740126 k    &   1251514 k   &   511388 k    &   815621 k   &   1241514 k   &   425893 k\\ \hline
\textbf{Energy} & 31569 µJ  &   49661 µJ    &   18092 µJ     &   59210 µJ    &   100121 µJ   &   40911 µJ    &   65250 µJ    &   99321 µJ    &   34071 µJ\\ \hline

            & \multicolumn{3}{c|}{\textbf{512, 512, 32 / 256, 256, 32}} & \multicolumn{3}{c|}{\textbf{512, 512, 32 / 256, 256, 64}} & \multicolumn{3}{c|}{\textbf{512, 512, 64 / 256, 256, 32}} \\ \hline
            
%               & MCC           &   Baseline    &   Saving      &   MCC         &   Baseline    &   Saving      &   MCC         &   Baseline    &   Saving\\ \hline
\textbf{MAC}    & 1516712 k &   2483028 k   &   966316 k    &   1577894 k   &   2483028 k   &   905134 k    &   295946846 k   &   4966056 k   &  2006587 k\\ \hline
\textbf{Energy} & 121337 µJ &   198642 µJ   &   77305 µJ    &  126231 µJ  &   198642 µJ   &   72411 µJ    &   236757 µJ   &   397284 µJ   &  160527 µJ\\ \hline

\end{tabular}
\label{table:EnergyPerInference}
\end{table*}

\begin{figure*}
    \centering
%    \begin{subfigure}
        \centering
        \includegraphics[width=2.8in]{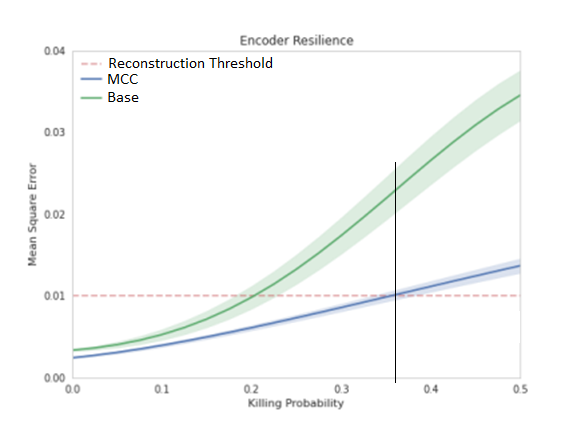}
%        \subcaption{}
%    \end{subfigure}%
%    \begin{subfigure}
        %\includegraphics[width=1\textwidth]{Figures/DeepMutualInformation.pdf}
        \includegraphics[width=2.6in]{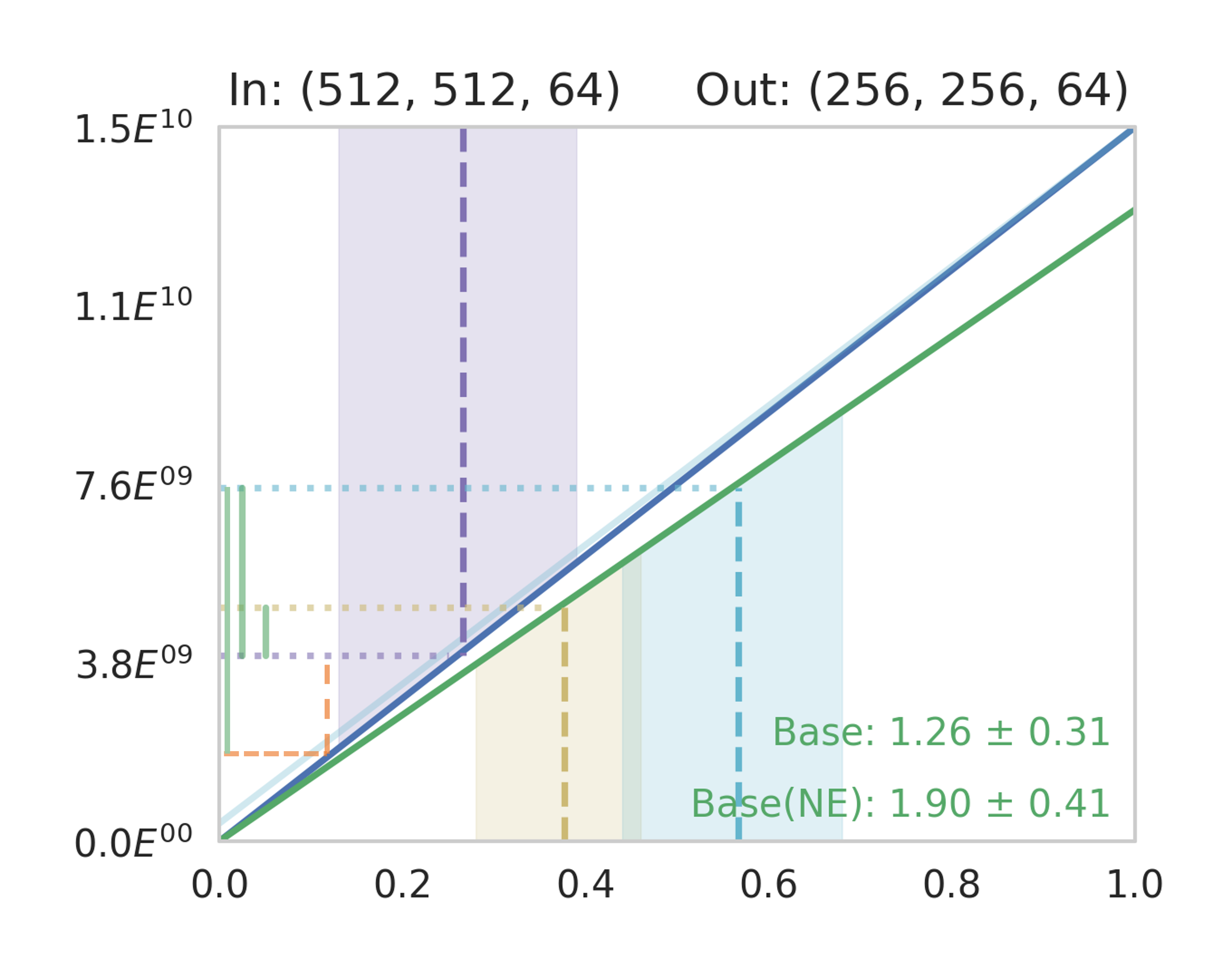}
%        \subcaption{}
%    \end{subfigure}
        \caption{(a) Random killing of up to 36\% cells in MCC could still achieve good accuracy with only 12.8\% overall neural activity (left) (b) MCC saves up to $245759\mu J$ energy per inference (62\% better than the baseline) (right).}
    \label{fig:deep_multimodal_representation}
\end{figure*}
\begin{figure*}
    \centering
%    \begin{subfigure}
        \centering
        \includegraphics[width=3.3in]{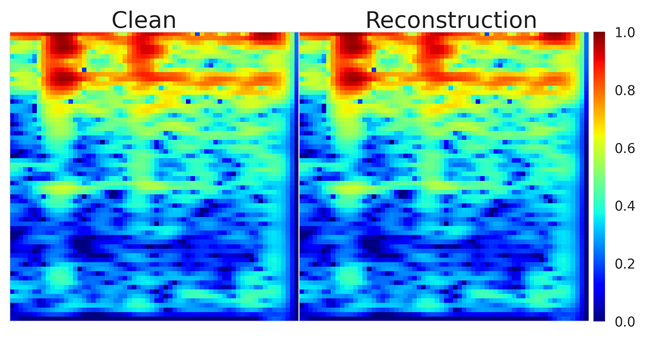}
%        \subcaption{}
%    \end{subfigure}
%    \begin{subfigure}
        %\includegraphics[width=1\textwidth]{Figures/DeepMutualInformation.pdf}
        \includegraphics[width=3.3in]{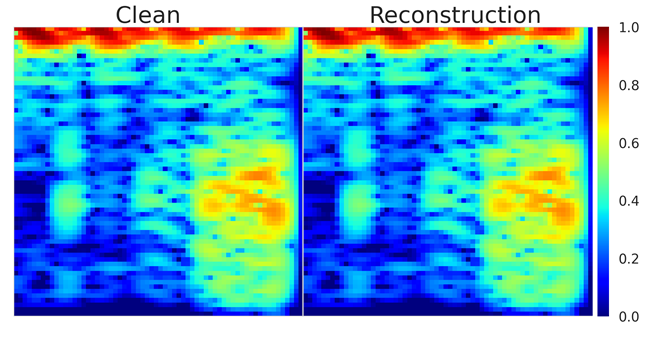}
%        \subcaption{}
%    \end{subfigure}
    \caption{STFT reconstruction (training): (a) MCC (left) (b) Baseline (right).}
    \label{fig:deep_mi_task}
\end{figure*}

\begin{figure*}
    \centering
%    \begin{subfigure}
        \centering
        \includegraphics[width=1\textwidth]{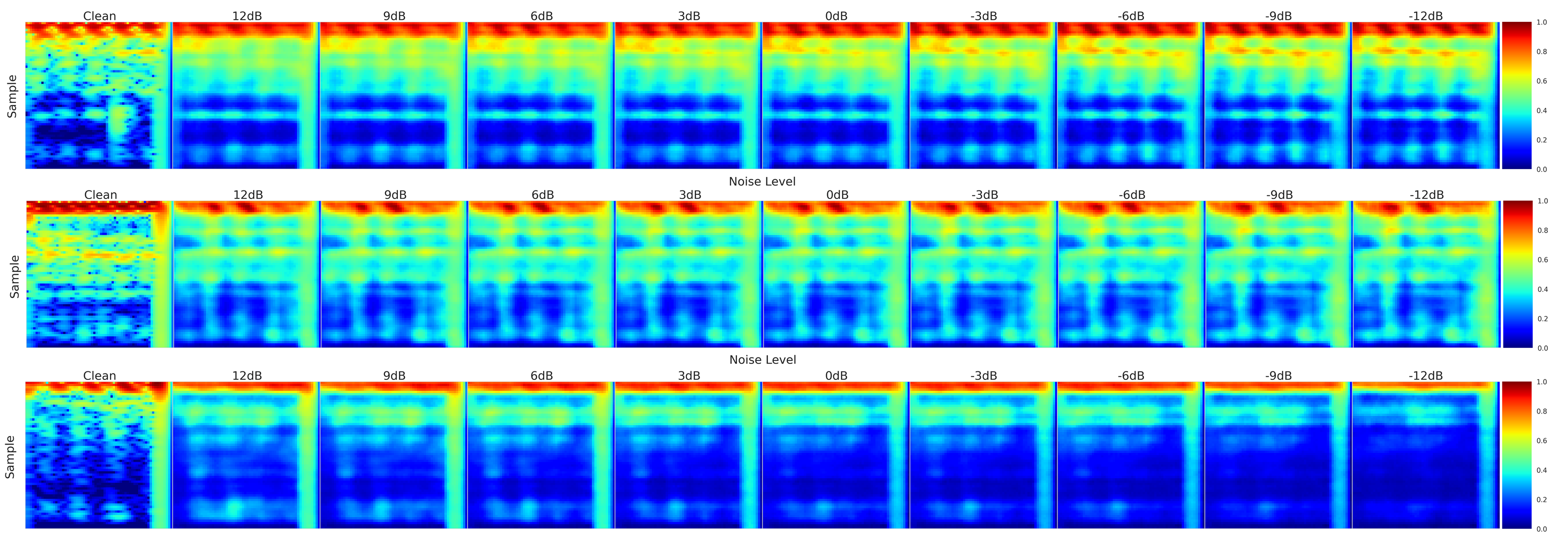}
%        \subcaption{MCC}
%    \end{subfigure} \\
%    \begin{subfigure}
        %\includegraphics[width=1\textwidth]{Figures/DeepMutualInformation.pdf}
        \includegraphics[width=1\textwidth]{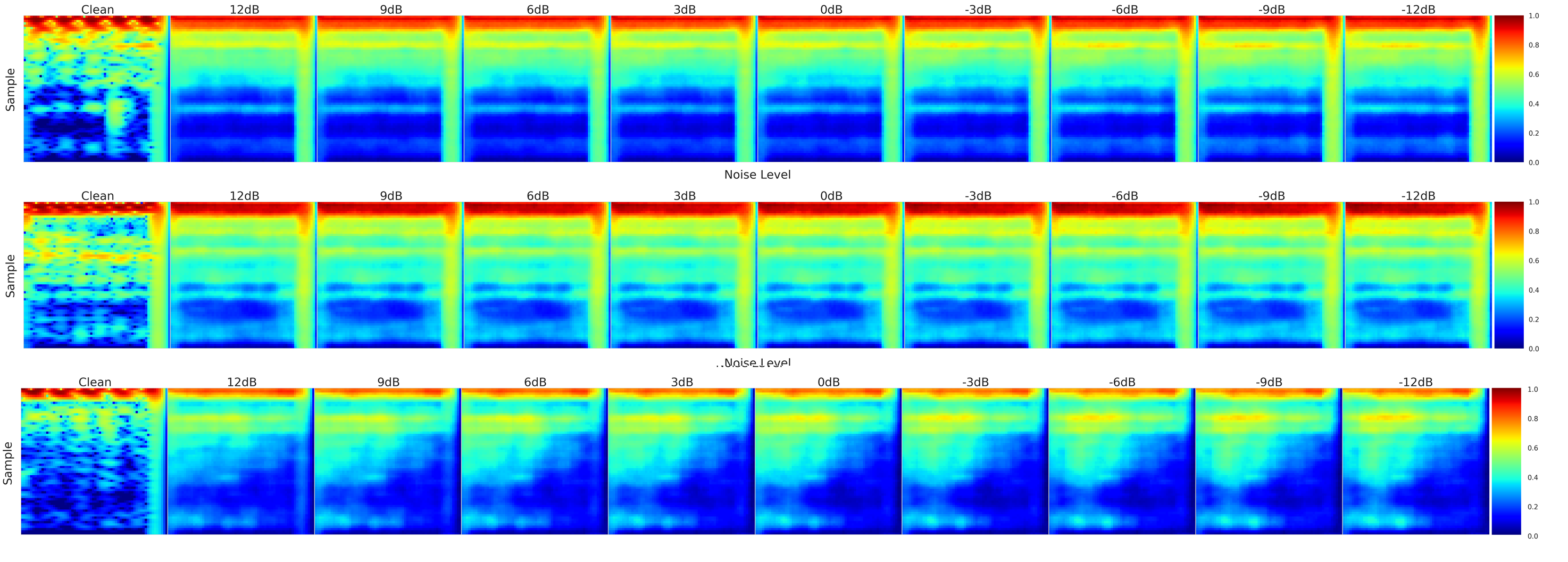}
%        \subcaption{Baseline}
%    \end{subfigure}
    \caption{Generalization/ testing: MCC vs. Baseline STFT reconstruction}
    \label{fig:deep_mi_task}
\end{figure*}

\begin{figure} 
	\centering
	\includegraphics[trim=0cm 0cm 0cm 0cm, clip=true, width=0.5\textwidth]{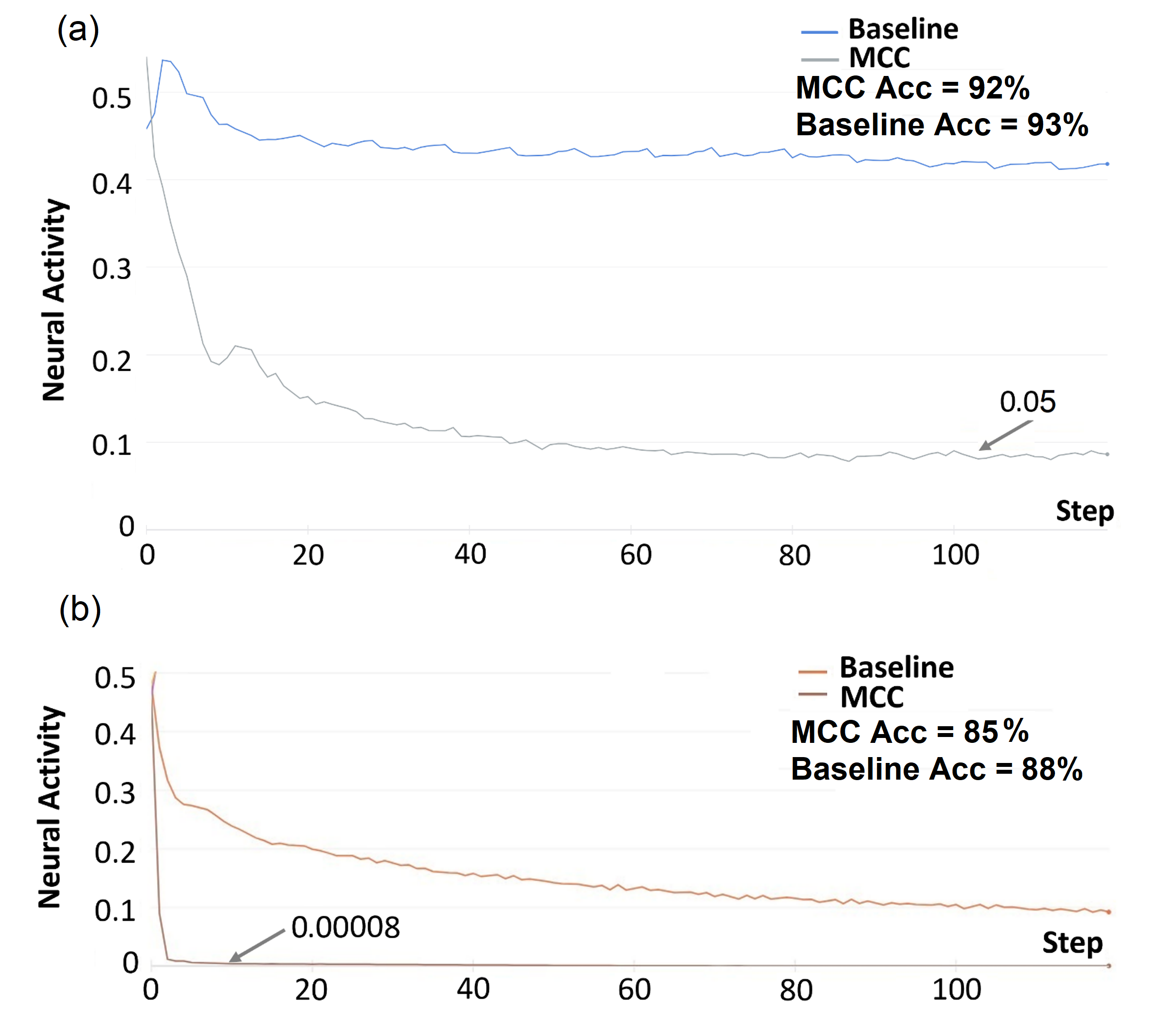}
	\caption{Supervised training MCC vs. Baseline: (a) 20 million parameters (b) 44 million parameters \cite{adeel2022context}.}
	\label{l5pc}
 \vspace{-1em}
\end{figure}

To estimate FPGA resources, a shallow multimodal model is first implemented with the network structure of $X^a_t$=22i:24i:12h:6h:22o for audio stream and $X^v_t$=50i:24h:12h:6h:22o for video stream. Measured energy values and true resource utilisation are reported for shallow models (Tables I and II) given the capacity of the available FPGA, whereas based on the used resources, an estimated energy consumption is reported for deep models (Figs. 6-10 and Table III). Fig. 5(a) depicts the training performance of the shallow model for semi-supervised AV speech processing. It can be seen that MCC quickly converges to the high MI as compared to the baseline models. However, MINE with concatenation \cite{Poria2017}\cite{kose2021multimodal} outperforms the standard MINE \cite{belghazi2018mutual}, and MINE with attention (baseline) \cite{bhatti2021attentive} outperforms the MINE with concatenation model. This learning trend aligns with the empirical Gaussian random variables dataset as shown in Fig. 5(b). MCC's remarkable performance improvement is due to its reduced neural activity property that enables the network to identify the most relevant features at very early stages in the network, avoiding transmitting irrelevant information to the higher network layers. As shown in Table 1, MCC achieves the minimum MSE with only 20\% neural activity consuming only 0.184$\mu J$ as compared to the baseline that has relatively high MSE consuming 4.13X more energy and 3.45X more processing time. Similarly, MCC consumes approx. half of the hardware resources as compared to the baseline model as shown in Table 2. 

Deep learning results reflecting the same trends for supervised clean-speech signal reconstruction. Deep MCC converges faster than the baselines (Fig. 6a) with only 20\% overall neural activity during training (Fig. 6b). It is to be noted that MCC learns at very early stages in the network what is relevant and what is not, thus, only neurons that transmit relevant information are active. The corresponding MAC operations are summarised in Fig. 7 and Table 3. It is to be noted that MCC could save up to 160527 $\mu$J of energy per inference i.e., 40\% less than the baseline model. During training, this energy-saving could be multiplied by the number of training updates e.g., 50K X 160527 $\mu$J. Furthermore, given the remarkable resilience property of MCC (considering sparsity) as shown in Fig. 8a, the energy-saving reaches up to 245759 $\mu$J per inference i.e., 62\% less than the baseline model as shown in Fig. 8b. In training, this could be multiplied by the number of training updates e.g., 50K X 245759 $\mu$J. Figs. 9 and 10 depict the clean STFT reconstruction for training and testing samples. It is observed that both MCC and baseline perform equally well for training samples, whereas MCC outperforms the baseline model in testing. 
For energy consumption estimations, a fixed-point was used to solve the issue of large resource utilization. This is because the mantissa defining the fractional value is suitably accurate even at low bit-width. Feature maps, biases, and weights were reduced from 32-bit floating points to 11-bit fixed points (Q3.7) using the data width quantization technique. It was observed that MSE increases drastically when the data width is smaller than 11-bit, while the performance is maintained when the data width is larger than or equal to 11-bit which was used for the hardware implementation. \\
When applied to solve a supervised learning problem, MCC is shown to drop an overall neural activity to 0.05 compared to 0.44 in the baseline (Figure 11) \cite{adeel2022context}. It is worth mentioning that neurons in MCC evolve quickly and reach this low neural activity in just a few training updates which further increases the efficiency. For a larger model comprising 44 million parameters, the neural activity reduces to less than 0.008\% i.e., ~1250x less (per FF transmission) than the baseline. However, this comes at the cost of reduced reconstruction accuracy for MCC (85\%) and baseline (88\%), respectively. Future work includes tuning and optimisation of MCC to search for Pareto-optimal.
\section{Conclusion}
In this paper, we presented a novel highly-distributed parallel implementation of our brain-inspired, non-von Neumann MCC architecture on a Xilinx UltraScale+ MPSoC device. The hardware architecture is evaluated using a benchmark AV speech enhancement problem, and exploits a cognitively-inspired, context-sensitive two-point L5PC neuron that quickly evolves during training and becomes highly selective in processing only the most salient data, instead of processing everything. This enables individual neurons to activate only when the received information is relevant to the task at hand. Our proposed hardware architecture emulates this cognitive behaviour by not propagating a synaptic signal of value zero in the network, which, in turn, avoids dynamic power consumption. This property is posited to be very useful for on-chip training and testing of both shallow and DNNs in future neuromorphic cognitive systems. For shallow models, the MCC has been shown in our pilot experiments to achieve 62\% better accuracy with 4.13X less energy consumption and 3.45X less processing time. For deep models with no sparsity, the MCC is seen to be 40\% more energy-efficient compared to the baseline and could save up to 160527 $\mu$J energy per inference during testing and 160527 $\times$ 50K $\mu$J during training. Considering sparsity, the MCC is 62\% more energy-efficient compared to the baseline and could save up to 245759 $\mu$J energy per inference during testing and 245759 $\times$ 50k $\mu$J during training. Similarly, for supervised training, the energy saving can potentially reach up to epochs$\times$1250x but at the cost of reduced accuracy \cite{adeel2022context}. The ongoing work involves evaluating different modulatory transfer functions to achieve better energy-accuracy trade-off. Certainly, the energy-saving per inference during testing could be multiplied with the number of inferences when the models are practically deployed. Our ongoing work includes implementing supervised training with MCC on MPSoC device. 

It is worth mentioning that this is the first time the two-point L5PC has been shown to provide useful energy-efficient computation at this scale, despite its discovery in 1999 \cite{larkum1999new} and theoretical predictions of it prior to that \cite{phillips1995discovery}\cite{kay1997activation}\cite{kay1998contextually}. Our MCC based neuromorphic model is more directly inspired by neuroscience and psychology compared to existing deep learning algorithms. In particular, the MCC is supported by recent neurobiological studies \cite{aru2020cellular}\cite{bachmann2020dendritic}\cite{shin2021memories}\cite{benjamin2021neocortical}\cite{shine2016dynamics}\cite{shine2019human}\cite{shine2019neuromodulatory}\cite{shine2021computational}\cite{marvan2021apical}, and is inherently energy-efficient. It does not require any special hardware design compared to other sparsity techniques \cite{chen2018sparse}\cite{hoefler2021sparsity}\cite{makhzani2015winner}\cite{kurtz2020inducing}\cite{mocanu2018scalable}\cite{ahmad2019can} \cite{changpinyo2017power}\cite{gale2020sparse}. The latter are difficult to exploit on modern hardware technology that is typically designed for regular dense data structures. Recently, a few approaches such as \cite{hunter2021two} have shown lower resource utilisation based on complementary kernel sparsity, however their application to real-world big data problems is yet to be demonstrated. 

We hypothesise that the proposed approach can be a step-change in understanding the brain's mysterious energy-saving mechanism. This, in turn, could pave the way to address multiple challenges and constraints  associated with adaptive design and real-time on-chip implementation of future multimodal technologies, such as audio-visual hearing-assistive devices \cite{adeel2021contextual}. The latter will require optimising a range of required tradeoffs including preservation of privacy, latency, energy, and speech intelligibility. In contrast, the MCC can potentially process everything on a single device (ESD) instead of on the Cloud \cite{adeel2020novel} or Edge \cite{li2019edge}. Ongoing work includes developing more compact MCC architectures and their integration with spiking neurons. In addition, new adaptive hardware architectures are being explored that can leverage the MCC's precisely controlled firing property to further reduce and optimise their energy consumption, latency and memory requirements for challenging real-world applications.

\section{Acknowledgments}
This research was supported by the UK Engineering and Physical Sciences Research Council (EPSRC) Grant Ref. EP/T021063/1. We would like to acknowledge Dr James Kay from the University of Glasgow and Professor Newton Howard from the Oxford Computational Neuroscience Lab for their advice and support, including reviewing of our work, appreciation, motivation, and encouragement. 

\section{Contributions}
AA conceived, developed, and simulated the original idea, wrote the manuscript, and analysed the results. AA, AA2, and KA performed the simulations and analysed the results. AA and WAP provided the psychoneuroscientific inspiration and advised on terminology and presentation. AA and AH provided the cognitive AV assistive technology inspiration. AA, AA2, and TA advised on practical implementation of AI algorithms on hardware.

% MF, MR, and KA improved the literature review and concept, designed the experiments, and performed the simulations.
% \bibliographystyle{unsrt}

%, and  Prof. Newton Howard from Oxford Computational Neuroscience
\section{Competing interests}
AA has a provisional patent application for the algorithm described in this article. The other authors declare no competing interests.
\bibliographystyle{IEEEtran}
\bibliography{referdif1.bib}

\end{document}